\documentclass{article}

\PassOptionsToPackage{numbers, compress}{natbib}

\usepackage{neurips_2026}

\usepackage[utf8]{inputenc} 
\usepackage[T1]{fontenc}    
\usepackage{hyperref}       
\usepackage{url}            
\usepackage{booktabs}       
\usepackage{amsfonts}       
\usepackage{nicefrac}       
\usepackage{microtype}      
\usepackage{xcolor}         

\usepackage{amsthm}
\usepackage{amsmath}
\usepackage{graphicx}
\usepackage{float}

\newtheorem{theorem}{Theorem}[section]

\newtheorem{lemma}[theorem]{Lemma}
\newtheorem{definition}[theorem]{Definition}

\usepackage{algorithm}
\usepackage{algorithmic}
\usepackage{tikz}
\usetikzlibrary{calc,intersections}
\usetikzlibrary{shapes, fit}
\usepackage{tikz-cd}

\usetikzlibrary{arrows.meta,positioning,fit,calc}

\tikzset{
  box/.style = {draw, rounded corners=6pt, minimum width=36mm, minimum height=12mm},
  bigbox/.style = {draw, rounded corners=6pt, minimum width=48mm, minimum height=12mm},
  tablebox/.style = {draw, minimum width=78mm, minimum height=12mm},
  dotprod/.style = {draw, circle, inner sep=1pt, minimum size=6mm},
  >={Stealth[length=2mm,width=2mm]}
}

\title{High Entropy Regularization Leads to Symmetry-Equivariant Policies in Dec-POMDPs}

\author{
  Johannes Forkel$^{1}$, Constantin Ruhdorfer$^{2}$, Michael Beukman$^{1}$\\\textbf{Andreas Bulling}$^{2}$, \textbf{Jakob Foerster}$^{1}$ \\
  $^{1}$FLAIR, Department of Engineering Science, University of Oxford, United Kingdom \\
  $^{2}$Collaborative Artificial Intelligence, University of Stuttgart, Germany
}

\begin{document}

\maketitle
\vspace{-0.5cm}
\begin{abstract}
    We prove that in any Dec-POMDP, sufficiently high entropy regularization ensures that the policy gradient flow with tabular softmax parametrization always converges, for any initialization, to the same joint policy, and that this joint policy is equivariant w.r.t. all symmetries of the Dec-POMDP. In particular, policies coming from different initializations will be fully compatible, in that their cross-play returns are equal to their self-play returns. Through extensive evaluation of independent PPO, arguably the standard baseline deep multi-agent policy gradient algorithm, in the Hanabi, Overcooked and Yokai environments, we find that the entropy coefficient has a massive influence on the cross-play returns between independently trained policies, and that the decrease in self-play returns coming from increased entropy regularization can often be counteracted by greedifying the learned policies after training. In Hanabi in particular we achieve a new SOTA in inter-seed cross-play this way. While we give examples of Dec-POMDPs in which one cannot learn the optimal symmetry-equivariant policy this way, both our theoretical and empirical results suggest that one should consider far higher entropy coefficients during hyperparameter sweeps in Dec-POMDPs than is typically done. Code for our experiments can be found at \url{https://github.com/jforkel/JAX-OBL}.
\end{abstract}

\section{Introduction}
Training agents that can coordinate well with novel partners is a central problem in cooperative multi-agent reinforcement learning (MARL). When using \textit{self-play} (SP) to train a joint policy in a decentralized partially observable Markov decision process (Dec-POMDP), this can result in the policy adopting a coordination convention which breaks symmetries of the Dec-POMDP. When such a policy is then paired in \textit{cross-play} (XP) with another policy which was not seen during training, e.g. with humans, or a policy trained with a different algorithm or the same algorithm but a different random seed, this can result in coordination failures, since the coordination conventions of the different policies are incompatible with each other. For an example of such incompatible symmetry-breaking, consider the one-round simultaneous-action game with common payoff matrix
{\setlength{\abovedisplayskip}{3pt}
\setlength{\belowdisplayskip}{3pt}
\begin{equation} \label{eq:payoff}
    \begin{bmatrix}
        2 & -2 & 1 \\
        -2 & 2 & 1 \\
        1 & 1 & 1
    \end{bmatrix}.
\end{equation}}
In this Dec-POMDP both joint policies which are optimal in SP break the symmetry in opposing ways: either both local policies always choose the first action, or both always choose the second action, for an SP return of $2$. Those two joint policies are incompatible though, in that their XP return is $-2$: when a local policy from one of those joint policies is paired with a local policy of the other, then this gives a return of $-2$. Given that one cannot deduce a preference for either one of those joint policies from the Dec-POMDP, the average XP return in a large population of SP optimal policies will be $0$. This is lower than $1$, which is the average XP return between any number of optimal symmetry-equivariant joint policies, under which both local policies always choose the third action.

In large and complex Dec-POMDPs like Hanabi \cite{bard2020hanabi} and Overcooked \cite{carroll2019utility}, standard MARL algorithms like independent PPO (IPPO) \cite{witt2020} or independent DQN have been reported to produce joint policies which, while achieving high returns in SP, do very poorly when paired in XP between independent seeds of the same algorithm \cite{bard2020hanabi, hu2020other, muglich2022equivariant, muglich2025expectedreturnsymmetries, gessler2025}, and in XP with human-proxy bots \cite{hu2021off, dizdarevic2025}. This is because those algorithms are unconstrained in which of the many different conventions they can converge to, and thus mostly converge to conventions which break symmetries of the Dec-POMDP. Many highly specialized algorithms were designed to produce policies which do not break symmetries and are compatible with each other in XP, e.g. \cite{hu2020other, cui2021k, hu2021off, cui2022offteamlearning, lupu2021trajectory, muglich2022equivariant, muglich2025expectedreturnsymmetries, lauffer2025robustdiversemultiagentlearning}. 

\paragraph{Our theoretical contribution:} We prove that with sufficiently high entropy regularization, the flow of the tabular softmax policy gradient will always converge, for any initialization, to a unique joint policy which is equivariant w.r.t. all symmetries of the Dec-POMDP. Thus, for sufficiently high entropy regularization, policies from different initializations will be fully compatible with each other, in that their SP returns are equal to their XP returns. We also show that a similar result does \textit{not} hold when using instead the gradient of the maximum entropy RL objective: we construct a simple counterexample where no matter how high the entropy coefficient is, the maximum entropy RL objective will have two incompatible maxima that policy gradient ascent can converge to.

\paragraph{Our empirical contribution:} We find that our theoretical predictions carry over into practice, where the drop in SP returns coming from high entropy regularization can often be counteracted by greedifying the policies after training. Using standard IPPO implementations in Hanabi and Overcooked, with all hyperparameters except the entropy coefficient being fairly standard (see Appendix \ref{appendix:hyperparameters}), we achieve very high returns in inter-seed XP as we increase the entropy regularization, with SP and XP being almost equal. In Hanabi in particular we find that vanilla IPPO with an LSTM architecture and an entropy coefficient of $0.05$ achieves close to perfect inter-seed XP, surpassing all previous highly specialized algorithms \cite{hu2020other, cui2021k, hu2021off, cui2022offteamlearning, lupu2021trajectory, muglich2022equivariant, muglich2025expectedreturnsymmetries}. In Overcooked, we show that XP of the greedified policies can improve even as one increases the entropy coefficient to the extremely large value $0.50$. For reference, in previous works on Overcooked that we are aware of, e.g. \cite{carroll2019utility, gessler2025, jha2025, lauffer2025robustdiversemultiagentlearning}, the IPPO entropy coefficient is at most $0.1$ and sometimes as low as $0.005$. While the recipe of high entropy regularization during training and greedification after training will produce symmetry-equivariant policies, the resulting policies might be very poor in SP, even among the symmetry-equivariant policies. We show in a toy game, as well as in the recently published Yokai learning environment \cite{ruhdorfer2025yokailearningenvironmenttracking}, that when the entropy regularization is high enough to ensure that all seeds converge to the same policy, that the resulting policy, even when greedified, is extremely poor, and in the case of the toy environment provably suboptimal among the symmetry-equivariant policies. Despite this obvious limitation, our results imply that when the aim is to learn a symmetry-equivariant policy in a Dec-POMDP, e.g. in order to be able to coordinate with previously unseen other policies, then during hyperparameter sweeps for policy gradient methods one should measure both SP and XP and sweep the entropy coefficient over a far bigger range than is typically done.

\section{Background}

\paragraph{Decentralized Partially Observable Markov Decision Processes (Dec-POMDPs):} We formalize the cooperative multi-agent setting as a Dec-POMDP, where we use notation from \cite{ravindran2003} to slightly generalize the definition from \cite{oliehoek2016} to account for different action spaces in different states:

\begin{definition}
A Dec-POMDP is defined as an 11-tuple $(\mathcal{S}, n, \{\mathcal{A}^i\}_{i=1}^n, \{\mathcal{O}^i\}_{i=1}^n, \{ \Psi^i \}_{i = 1}^n, \mathcal{T}, \mathcal{R}, \{\mathcal{U}^i\}_{i=1}^n, T, \gamma, b_0)$. $\mathcal{S}$ is the finite state space, $b_0$ is the initial state distribution, and $n$ is the number of agents. $\mathcal{A}^i$ and $\mathcal{O}^i$ are the finite local action and observation spaces for agent $i$, and  $\mathcal{A} = \prod_{i = 1}^n \mathcal{A}^i$,  $\mathcal{O} = \prod_{i = 1}^n \mathcal{O}^i$ are the joint action and observation spaces. $o_{t}^i := \mathcal{U}^i(s_{t}) \in \mathcal{O}^i$ is the local observation agent $i$ receives in state $s_t$, and we set $\mathcal{U}(s_{t}) := (o_{t}^1, ..., o_{t}^n) \in \mathcal{O}$. $\Psi^i \subseteq \mathcal{O}^i \times \mathcal{A}^i$ is the set of admissible local observation-action pairs, and $\mathcal{A}^i(o^i) := \{ a^i \in \mathcal{A}^i | (o^i, a^i) \in \Psi^i \}$ is the space of legal local actions of agent $i$ given $o^i$. We set $\mathcal{A}(s) := \prod_{i = 1}^n \mathcal{A}^i(\mathcal{U}^i(s))$ to be the space of legal joint actions in state $s$. $\mathcal{T}(s_{t+1} | s_t, a_t)$ is the probability to transition to state $s_{t+1}$ when taking the legal joint action $a_t = (a_t^1, ..., a_t^n) \in \mathcal{A}(s_t)$ in state $s_t$. The rewards are $r_{t+1} = \mathcal{R}(s_{t+1}, s_t, a_t) \in \mathbb{R}$, where $a_t \in \mathcal{A}(s_t)$. $T \in \mathbb{N}$ is the horizon, i.e. $s_T$ is always a terminal state and $\gamma \in (0, 1]$ is the discount factor. Given $t \in \{0, ..., T\}$, the state-action history (SAH) is given by $\tau_t = (s_0, a_0, ..., s_{t-1}, a_{t-1}, s_t)$, and for each $i \in \{1, ..., n\}$ the local action-observation history (AOH) of agent $i$ is given by $\tau_t^i = (o_0^i, a_0^i, o_1^i, \dots, a_{t-1}^i, o_t^i)$. For every $i$, $\tau_t^i$ is a deterministic function of $\tau_t$, and a joint policy $\pi = (\pi^1, \dots, \pi^n)$, consisting of local policies $\pi^1, ..., \pi^n$, chooses a legal joint action $a_t = (a_t^1, ..., a_t^n) \in \mathcal{A}(s_t)$, with probability $\pi(a_t | \tau_t) := \prod_{i=1}^n \pi^i(a_t^i | \tau_t^i)$. We denote the set of joint policies by $\Pi$, and define the self-play (SP) objective $J_{\text{SP}}: \Pi \rightarrow \mathbb{R}$ as the expected return: $J_{\text{SP}} (\pi) = \mathbb{E}_{\tau_T \sim \pi} \left[ \sum_{t = 0}^{T-1} \gamma^t \mathcal{R}(s_{t+1}, s_t, a_t) \right]$, where $p_{\pi_\theta}(\tau_T) := b_0(s_0) \prod_{t = 0}^{T-1} \pi_\theta(a_t | \tau_t) \mathcal{T}(s_{t+1} | s_t, a_t)$.
\end{definition}

Given a joint policy $\pi$, we define its \textbf{greedification} $\hat{\pi}$ by
\begin{align*}
    \hat{\pi}(a_t | \tau_t) := \begin{cases}
        \frac{1}{K} & \text{if } a_t \in \text{argmax}_{a \in \mathcal{A}(s_t)} \pi(a | \tau_t), \\
        0 & \text{else},
    \end{cases} = \begin{cases} 
        \frac{1}{K} & \text{if } \forall i: a_t^i \in \text{argmax}_{a^i \in \mathcal{A}^i(o_t^i)} \pi^i(a^i | \tau_t^i), \\
        0 & \text{else},
    \end{cases}
\end{align*}
where $K := | \text{argmax}_{a \in \mathcal{A}(s_t)} \pi(a | \tau_t)| = \prod_{i = 1}^n | \text{argmax}_{a^i \in \mathcal{A}^i(o_t^i)} \pi^i(a^i | \tau_t^i)|$ is the number of joint actions with the highest probability. In practice, greedification will almost always result in a deterministic policy, since no two actions will have exactly the same probability.

\paragraph{Zero-Shot Coordination (ZSC):} There are typically many joint policies in a Dec-POMDP which maximize the SP return $J_{\text{SP}}$, but those joint policies might be incompatible with each other. The XP return $J_{\text{XP}}$, defined as follows, measures the compatibility between different joint policies:

\begin{definition}[Cross-Play (XP)]
    Given a Dec-POMDP with $n$ players, we define the cross-play (XP) return $J_{\text{XP}}: \Pi^n \rightarrow \mathbb{R}$, between $n$ joint policies $\pi_1$, ..., $\pi_n$ by
    \begin{align*}
        J_{\text{XP}}(\pi_1, ..., \pi_n) := \mathbb{E}_{\phi \sim \text{Perm}(n)} \left[ J_{\text{SP}}((\pi_{\phi(1)}^1, ..., \pi_{\phi(n)}^n ))\right],
    \end{align*}
    where $\text{Perm}(n)$ is the set of permutations of $\{1, ..., n\}$.
\end{definition}

For example, for $n=2$, this becomes $J_{\text{XP}}(\pi_1, \pi_2) = \frac{1}{2} [J_{\text{SP}}((\pi_1^1, \pi_2^2)) + J_{\text{SP}}((\pi_2^1, \pi_1^2))]$. Then, given joint policies $\pi_1, ..., \pi_m$, we refer to the matrix $\left( J_{\text{SP}}((\pi_j^1, \pi_k^2)) \right)_{j,k = 1}^{m}$ as the XP matrix of $\pi_1, ..., \pi_m$.

The goal of zero-shot coordination (ZSC), introduced in \cite{hu2020other}, is to find algorithms which produce compatible joint policies across different implementations, in particular across different random seeds of the same implementation. In Hanabi, ZSC algorithms have led to large improvements compared to standard MARL algorithms, in XP returns between independently trained joint policies from different random seeds of the same implementation, and in XP with human-proxy bots \cite{hu2020other, cui2021k, hu2021off, cui2022offteamlearning, lupu2021trajectory, muglich2022equivariant, muglich2025expectedreturnsymmetries, dizdarevic2025}. As a minor contribution, we describe in Appendix \ref{appendix:XP estimators} a statistically correct way to estimate for a given algorithm the average XP and its standard error. This approach is different to what has been done in previous papers on ZSC, e.g. in the ones just cited.

\paragraph{Symmetries in Dec-POMDPs:} We generalize the definition of Dec-POMDP symmetries from \cite{hu2020other} to the case where the action space is not necessarily the same in every state:

\begin{definition}[Dec-POMDP Symmetries]
    A tuple $\phi:= (\phi_\mathcal{S}, (\phi_{\mathcal{A}^i})_{i = 1}^n, (\phi_{\mathcal{O}^i})_{i = 1}^n)$ of bijections
    \begin{align}
    \begin{split}
        \phi_{\mathcal{S}}: \mathcal{S} \rightarrow \mathcal{S}, \quad \phi_{\mathcal{A}^i}: \mathcal{A}^i \rightarrow \mathcal{A}^i, \, \, i = 1, ..., n, \quad \phi_{\mathcal{O}^i}: \mathcal{O}^i \rightarrow \mathcal{O}^i, \, \, i = 1, ..., n,
    \end{split}
    \end{align}
    is a Dec-POMDP symmetry if $\forall s, s' \in \mathcal{S}, \, \, a \in \mathcal{A}(s)$ it holds that $b_0(\phi_{\mathcal{S}}(s)) = b_0(s)$ and
    \begin{align}
    \begin{split}
        \mathcal{T}(s' | s, a) =& \mathcal{T} (\phi_{\mathcal{S}}(s') | \phi_{\mathcal{S}}(s), \phi_{\mathcal{A}}(a) ), \quad \phi_{\mathcal{O}} (\mathcal{U}(s)) = \mathcal{U} (\phi_{\mathcal{S}}(s)), \\
        \mathcal{R}(s', s, a) =& \mathcal{R} (\phi_{\mathcal{S}}(s'), \phi_{\mathcal{S}}(s), \phi_{\mathcal{A}}(a)), \quad \phi_{\mathcal{A}}(\mathcal{A}(s)) = \mathcal{A}(\phi_\mathcal{S}(s)),
    \end{split}
    \end{align}
    where for joint actions $a = (a^1, ...., a^n) \in \mathcal{A}(s)$, joint observations $o = (o^1, ..., o^n) \in \mathcal{O}$, and SAHs $\tau_t = (s_0, a_0, ..., a_{t-1}, s_t)$, we define $\phi_{\mathcal{A}}(a)$,  $\phi_{\mathcal{O}}(o)$, and $\phi(\tau_t)$ by applying $\phi_{\mathcal{S}}$, $\phi_{\mathcal{O}^i}$ and $\phi_{\mathcal{A}^i}$ elementwise. We denote by $\Phi$ the set of all Dec-POMDP symmetries of a given Dec-POMDP.

    Dec-POMDP symmetries act on joint policies in the following way: for $\phi \in \Phi$ and $\pi \in \Pi$, we define the joint policy $\phi(\pi) := (\phi(\pi^1), ..., \phi(\pi^n))$ by the formula $\phi(\pi^i)(a^i | \tau^i) := \pi^i(\phi^{-1}(a^i) | \phi^{-1}(\tau^i)).$
\end{definition}

We define a joint policy $\pi$ to be symmetry-equivariant if $\phi(\pi) = \pi$ for all $\phi \in \Phi$, else we call $\pi$ symmetry-breaking. In order to learn symmetry-equivariant policies, \cite{hu2020other} introduced $J_{\text{OP}}$:

\begin{definition}[Other-Play \cite{hu2020other}]
    The Other-Play (OP) objective $J_{\text{OP}}$ is defined as:
    \begin{align*}
        J_{\text{OP}}(\pi) := \mathbb{E}_{(\phi^1, ..., \phi^n) \sim \Phi^n} \left[ J_{\text{SP}}( (\phi^1(\pi^1), ..., \phi^n(\pi^n)))\right].
    \end{align*}
\end{definition}

Maximizing $J_{\text{OP}}$ instead of $J_{\text{SP}}$ leads to symmetry-equivariant policies. For example, in the game with payoff matrix (\ref{eq:payoff}), $\Phi = \{ \text{Id}, \phi_{1 \leftrightarrow 2} \}$ where $\phi_{1 \leftrightarrow 2}$ corresponds to permuting the first two actions. Then $\phi_{1 \leftrightarrow 2}(\pi_1) = \pi_2$ and $\phi_{1 \leftrightarrow 2}(\pi_3) = \pi_3$, where for $j = 1, 2, 3$, $\pi_j$ denotes the joint policy under which both agents always choose the $j$-th action. Then $J_{\text{SP}}(\pi_1) = J_{\text{SP}}(\pi_2) = 2 > 0 = J_{\text{OP}}(\pi_1) = J_{\text{OP}}(\pi_2)$, and $J_{\text{SP}}(\pi_3) = J_{\text{OP}}(\pi_3) = 1$. Thus, optimizing for $J_{\text{OP}}$ here yields the optimal symmetry-equivariant policy. For another very explicit example of a Dec-POMDP symmetry see Figure \ref{fig:XP_cat/dog} and Appendix \ref{appendix:cat/dog}.

The big drawback of the OP algorithm is that one needs to know the symmetries of the Dec-POMDP a priori. \cite{muglich2025expectedreturnsymmetries} provide an algorithm to learn symmetries, which can then be used in the OP objective, but those symmetries are just approximate and the algorithm in \cite{muglich2025expectedreturnsymmetries} does not guarantee to find all of the Dec-POMDP symmetries. Furthermore, even when knowing all the symmetries, there can be multiple incompatible OP optimal policies \cite{treutlein2021new}.

\paragraph{Entropy Regularized Multi-Agent Policy Gradients:} We denote the entropy of the joint action distribution $\pi( \cdot | \tau_t)$ by $\text{Ent}(\pi( \cdot | \tau_t)) := - \sum_{a \in \mathcal{A}(s_t)} \pi(a | \tau_t) \log \pi( a | \tau_t)$, and similarly for the local action distribution $\pi^i( \cdot | \tau_t^i)$. Let $\theta \mapsto \pi_\theta$ be a differentiable parametrization of the joint policy. For an entropy coefficient $\alpha \geq 0$, we then define the entropy regularized policy gradient as follows:
\begin{align}
    &\nabla_\theta^\alpha J_{\text{SP}}(\pi_\theta) := \nabla_\theta J_{\text{SP}}(\pi_\theta) + \alpha \mathbb{E}_{\tau_T \sim \pi_\theta} \left[ \nabla_\theta \sum_{t = 0}^{T-1} \gamma^t \text{Ent}(\pi_\theta( \cdot | \tau_t)) \right] \\
    =& \mathbb{E}_{\tau_T \sim \pi_\theta} \Bigg[ \sum_{t = 0}^{T-1} \gamma^t \sum_{i = 1}^n \Bigg( \alpha \nabla_\theta \text{Ent}(\pi_\theta^i(\cdot | \tau_t^i)) + \nabla_\theta \log \pi_\theta^i(a_{t}^i | \tau_{t}^i) \sum_{t' = t}^{T-1} \gamma^{t'-t} \mathcal{R}(s_{t'+1}, s_{t'}, a_{t'}) \Bigg) \Bigg], \nonumber
\end{align}
Note that $\nabla_\theta^\alpha J_{\text{SP}}(\pi_\theta)$ is in general \textbf{not} a gradient, so in particular not the same as the gradient $\nabla_\theta J_{\text{SP}}^\alpha(\pi_\theta)$ of the maximum entropy RL objective
\begin{align}
    J_{\text{SP}}^\alpha(\pi_\theta) :=& J_{\text{SP}}(\pi_\theta) + \alpha \mathbb{E}_{\tau_T \sim \pi_\theta} \left[ \sum_{t = 0}^{T-1} \gamma^t \text{Ent}(\pi_\theta( \cdot | \tau_t)) \right], \quad \text{because} \nonumber \\
    \nabla_\theta J_{\text{SP}}^\alpha(\pi_\theta) - \nabla_\theta^\alpha J_{\text{SP}}(\pi_\theta) =& \alpha \mathbb{E}_{\tau_T \sim \pi_\theta} \left[ \sum_{t = 0}^{T-1} \nabla_\theta \log \pi_\theta(a_t | \tau_t) \sum_{t' = t + 1}^{T-1} \gamma^{t'} \text{Ent}(\pi_\theta(\cdot | \tau_{t'})) \right]. \label{eq:max ent no gradient}
\end{align}

The entropy bonus in PPO \cite{schulman2017proximalpolicyoptimizationalgorithms}, translated to recurrent neural networks necessary for Dec-POMDPs, can be viewed as an estimate of the entropy part in $\nabla_\theta^\alpha J_{\text{SP}}(\pi_\theta)$. In \cite{yu2022surprising} it was found that IPPO is effective in learning joint policies with high $J_{\text{SP}}$ in Hanabi and other Dec-POMDPs. In e.g. \cite{muglich2025expectedreturnsymmetries, gessler2025} however it was reported that IPPO achieves very low inter-seed XP in Hanabi and Overcooked.

\section{Why high entropy regularization leads to symmetry-equivariant policies}

In this section we assume a centered tabular softmax parametrization: for every $\tau_t^i$ which is possible in the Dec-POMDP, $\theta^i(a_t^i | \tau_t^i)$ is the logit corresponding to agent $i$ taking local action $a_t^i \in \mathcal{A}^i(o_t^i)$ given $\tau_t^i$, and $d \in \mathbb{N}$ is the total number of such logits, and $\Theta$ is the space of centered logits:
\begin{align*}
    \pi_{\theta}^i(a_t^i | \tau_t^i) = \frac{\exp(\theta^i(a_t^i | \tau_t^i))}{\sum_{b_t^i \in \mathcal{A}^i(o_t^i)} \exp(\theta^i(b_t^i | \tau_t^i))}, \quad
    \Theta := \left\{ \theta \in \mathbb{R}^d: \forall i, \tau_t^i: \sum_{a_t^i \in \mathcal{A}^i(o_t^i)} \theta^i(a_t^i | \tau_t^i) = 0 \right\}.
\end{align*}

We let $\theta_0 \in \Theta$ be the initial parameters, and set
\begin{align} \label{eq:policy gradient ascent}
    \theta_{k+1} = \theta_k + \eta_k \nabla_\theta^\alpha J_{\text{SP}}(\pi_{\theta_k}), \quad k \in \mathbb{N}_0,
\end{align}
for step sizes $\eta_k > 0$, $k \in \mathbb{N}_0$, satisfying the Robbins-Monro conditions $\sum_{k = 0}^\infty \eta_k = \infty$, and $\sum_{k = 0}^\infty \eta_k^2 < \infty$. When initializing $\theta_0$ randomly and/or using noisy estimates of the gradients, and assuming convergence, there are typically many different limiting policies $\pi_{\theta_\infty}$. Intuitively, increasing the entropy coefficient $\alpha$ ``makes the objective function more concave''\footnote{We used quotation marks here to emphasize that in general (see Equation \ref{eq:max ent no gradient}) there is not actually an objective function whose gradient is given by the vector field $\theta \mapsto \nabla_\theta^\alpha J_{\text{SP}}(\pi_\theta)$.}, since entropy is a strictly concave function on the probability simplex, having as its unique maximizer the uniform distribution. For $\alpha \rightarrow \infty$, when assuming exact values of $\nabla_\theta^\alpha J_{\text{SP}}(\pi_\theta)$ or unbiased estimates thereof, we can expect the limiting policy $\pi_{\theta_\infty}$ to always exist and be the uniformly random policy, for any $\theta_0 \in \Theta$. This is because as $\alpha \rightarrow \infty$ the relative contribution of the term $\nabla_\theta J_{\text{SP}}(\pi_\theta)$ in $\nabla_\theta^\alpha J_{\text{SP}}(\pi_\theta)$ becomes negligible, and one is just following the vector field $\theta \mapsto \mathbb{E}_{\tau_T \sim \pi_\theta} \left[ \nabla_\theta \sum_{t = 0}^{T-1} \text{Ent}(\pi_\theta( \cdot | \tau_t)) \right]$, which always leads to $\pi_{\theta_\infty}$ being the uniformly random policy. We prove the following in Appendix \ref{appendix:proof}.

\begin{theorem} \label{thm:unique limit}
    Let $\theta \mapsto \pi_\theta$ be the tabular softmax parametrization in a Dec-POMDP. Then there exists a finite entropy threshold $\alpha' \in [0, \infty)$, such that for all $\alpha > \alpha'$ there exists a unique $\theta^\alpha \in \Theta$ at which the vector field $\theta \mapsto \nabla_\theta^\alpha J_{\text{SP}}(\pi_{\theta})$ equals zero. Furthermore, $\pi_{\theta^\alpha}$ is symmetry-equivariant, i.e. $\phi(\pi_{\theta^\alpha}) = \pi_{\theta^\alpha}$ for all $\phi \in \Phi$. Finally, for any $\theta_0 \in \Theta$ it holds that $\lim_{x \rightarrow \infty} \theta_x = \theta^\alpha$, where $x \mapsto \theta_x$ is the solution to the ODE $\frac{\text{d}\theta_x}{\text{d}x} = \nabla_\theta^\alpha J_{\text{SP}}(\pi_{\theta_x})$ starting at $\theta_0$.
\end{theorem}

When assuming convergence of policy gradient ascent with non-infinitesimal step sizes, i.e. the sequence (\ref{eq:policy gradient ascent}), then Theorem \ref{thm:unique limit} implies that for $\alpha > \alpha'$ the limit can only be $\theta^\alpha$. We conjecture that one can use standard stochastic approximation results from e.g. \cite{borkar2023}, to extend Theorem \ref{thm:unique limit} and prove that for any $\theta_0 \in \Theta$ the sequence (\ref{eq:policy gradient ascent}), with unbiased estimates of $\nabla_\theta^\alpha J_{\text{SP}}(\pi_{\theta_k})$ with finite variance, like REINFORCE-type estimators, almost surely converges to a limit for all $\alpha > 0$.

\begin{figure}[t]
    \centering
    \includegraphics[width=0.225\textwidth]{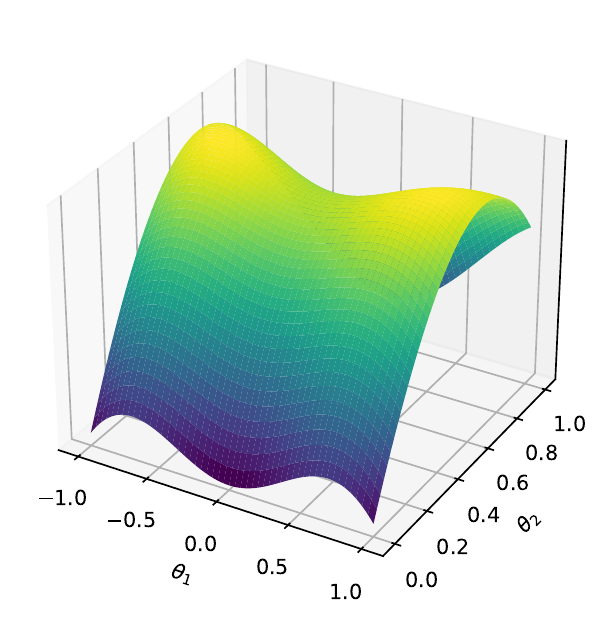}
    \includegraphics[width=0.225\textwidth]{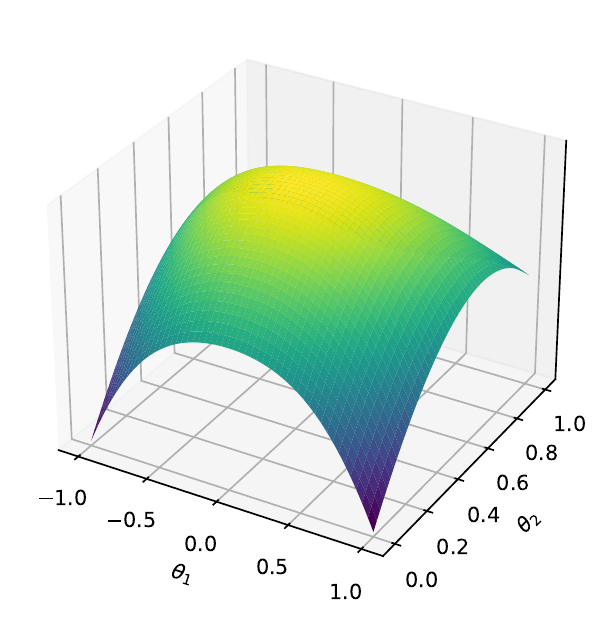}
    \includegraphics[width=0.23\textwidth]{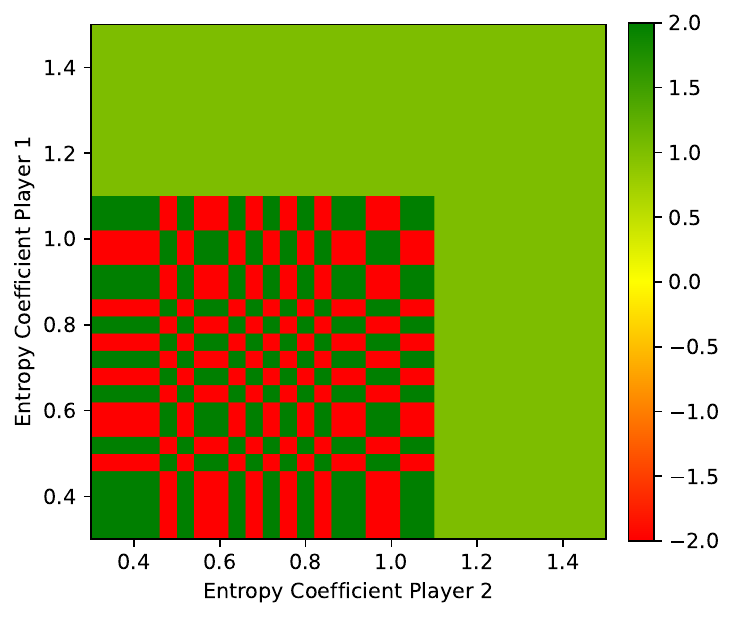}
    \includegraphics[width=0.21\textwidth]{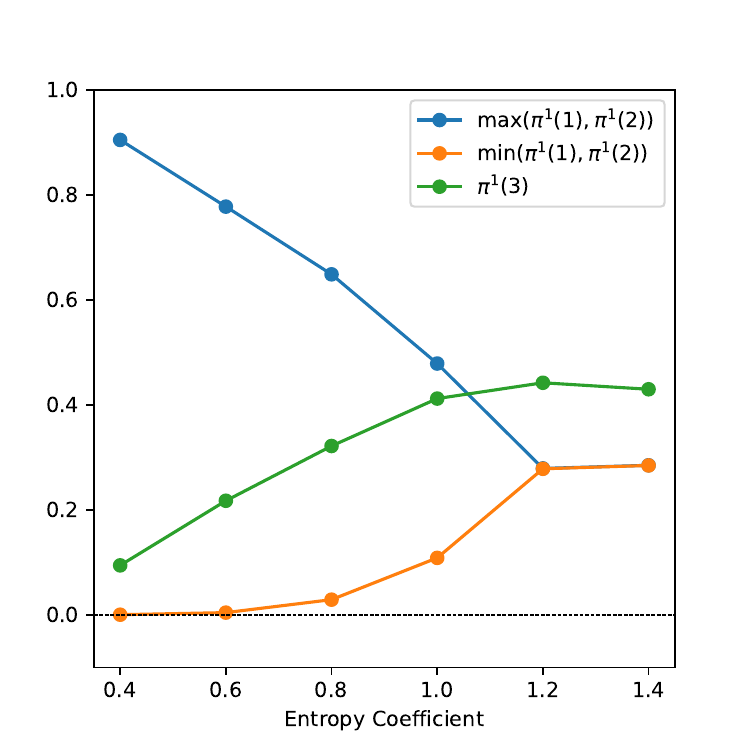}
    \caption{Consider the one-round simultaneous-action game with payoff matrix given in (\ref{eq:payoff}), and the parametrization $\theta = (\theta_1, \theta_2) \mapsto \text{softmax}(\theta_1, -\theta_1, \theta_2) = \pi^1_\theta(\cdot) = \pi^2_\theta(\cdot)$. \textbf{Left and second from left}: the function $\theta \mapsto J^\alpha_{\text{SP}}(\pi_\theta)$, with $\alpha = 1.0$ and $\alpha = 1.2$. \textbf{Second from right}: XP matrix between multiple greedified joint policies, which were trained with entropy regularized independent REINFORCE with baseline, with $\alpha \in \{0.4, 0.6, 0.8, 1.0, 1.2, 1.4\}$. 5 seeds per entropy coefficient. \textbf{Right}: action probabilities of one trained policy for each $\alpha$. \textbf{Takeaway}: In this game the entropy threshold $\alpha'$ lies between $1.0$ and $1.2$, i.e. for $\alpha > \alpha'$ the objective function $\theta \mapsto J_{\text{SP}}^\alpha(\pi_\theta)$ has a unique local maximum, the learned policies stop breaking the symmetry, and are fully compatible.}
    \label{fig:XP_levergame}
\end{figure}

\paragraph{Greedification after training:} When the entropy threshold $\alpha'$ is very high, then $\pi_{\theta^\alpha}$ might be very close to the uniformly random policy for $\alpha > \alpha'$, which means that $J_{\text{SP}}(\pi_{\theta^\alpha})$ might be very low. However, it is easy to see that the greedification $\hat{\pi}_{\theta^\alpha}$ is also symmetry-equivariant, and that $J_{\text{SP}}(\hat{\pi}_{\theta^\alpha})$ might be significantly larger than $J_{\text{SP}}(\pi_{\theta^\alpha})$. Regarding the latter: from Equation (\ref{eq:Q values}) in the proof of Theorem \ref{thm:unique limit} we see that if $\nabla_\theta^\alpha J_{\text{SP}}(\pi_\theta) = 0$ for a $\theta \in \Theta$, then $\theta^i(a_t^i | \tau_t^i) = \alpha^{-1} Q_{\pi_\theta}^i(a_t^i | \tau_t^i) - \text{Baseline}(\tau_t^i)$ for all $i, \tau_t^i, a_t^i$, where $Q_{\pi_\theta}^i(a_t^i | \tau_t^i) := \mathbb{E}_{\tau_T \sim \pi_\theta} \left[ \sum_{t' = t}^{T-1} \gamma^{t' - t} \mathcal{R}(s_{t'+1}, s_{t'}, a_{t'}) | a_t^i, \tau_t^i \right]$. Thus we see that if $\nabla_\theta^\alpha J_{\text{SP}}(\pi_\theta) = 0$, then $\hat{\pi}_\theta$ is greedy w.r.t. the local $Q$-values of $\pi_\theta$. Therefore it does not matter if $\pi_{\theta^\alpha}$ assigns a lot of probability to the suboptimal actions, so long as $\pi_{\theta^\alpha}$ identifies good actions by assigning them the highest $Q$-values. In practice, to find the entropy threshold and to see whether the learned policies identified good actions this way, one can train a pool of joint policies for each $\alpha$, until the greedified SP and XP in that pool are equal. For an example of this, see Figure \ref{fig:XP_levergame}.

\paragraph{Why this does not work with the maximum entropy RL objective:} Theorem \ref{thm:unique limit} is easily seen to be true in one-round Dec-POMDPs, i.e. when $T=1$, since then $\nabla_\theta^\alpha J_{\text{SP}}(\pi_\theta) = \nabla_\theta J_{\text{SP}}^\alpha(\pi_\theta)$ for all $\theta \in \mathbb{R}^d$, implying that then one is maximizing the actual objective function $\pi \mapsto J_{\text{SP}}^\alpha(\pi)$, which becomes strictly concave for large enough $\alpha$, and for which it is easily seen that $J_{\text{SP}}^\alpha(\phi(\pi)) = J_{\text{SP}}^\alpha(\pi)$ for all $\phi \in \Phi$. For an example of this phenomenon see again Figure \ref{fig:XP_levergame}. In general Dec-POMDPs however, when using $\nabla_\theta J_{\text{SP}}^\alpha(\pi_\theta)$ instead of $\nabla_\theta^\alpha J_{\text{SP}}(\pi_\theta)$, one can \textbf{not} guarantee that for large $\alpha$ the limiting policies will be the same for every initialization $\theta_0$. This is because when there are $2$ or more agents, the function $\pi \mapsto \mathbb{E}_{\tau_T \sim \pi} \left[ \sum_{t = 0}^{T-1} \text{Ent}(\pi( \cdot | \tau_t)) \right]$ can have multiple global maxima, and is thus in particular not necessarily concave. For an example, consider a modified version of the one-round game with payoff matrix (\ref{eq:payoff}), where if the reward in the first round is $2$, then there is an additional round where both agents can choose between $10$ dummy actions which all give a reward of $0$. Then there are two different policies which both maximize $\mathbb{E}_{\tau_T \sim \pi} \left[ \sum_{t = 0}^{T-1} \text{Ent}(\pi( \cdot | \tau_t)) \right]$, namely one where in the first round both agents mostly choose the first action, and one where both agents mostly choose the second action (in the second round both policies are uniformly random). Thus using $\nabla_\theta J_{\text{SP}}^\alpha(\pi_\theta)$ and increasing $\alpha$ in this game will \textit{increase} the incentive to break the symmetry.

\paragraph{An illustrative toy example:}

\begin{figure}
    \centering
    \includegraphics[width=0.40\textwidth]{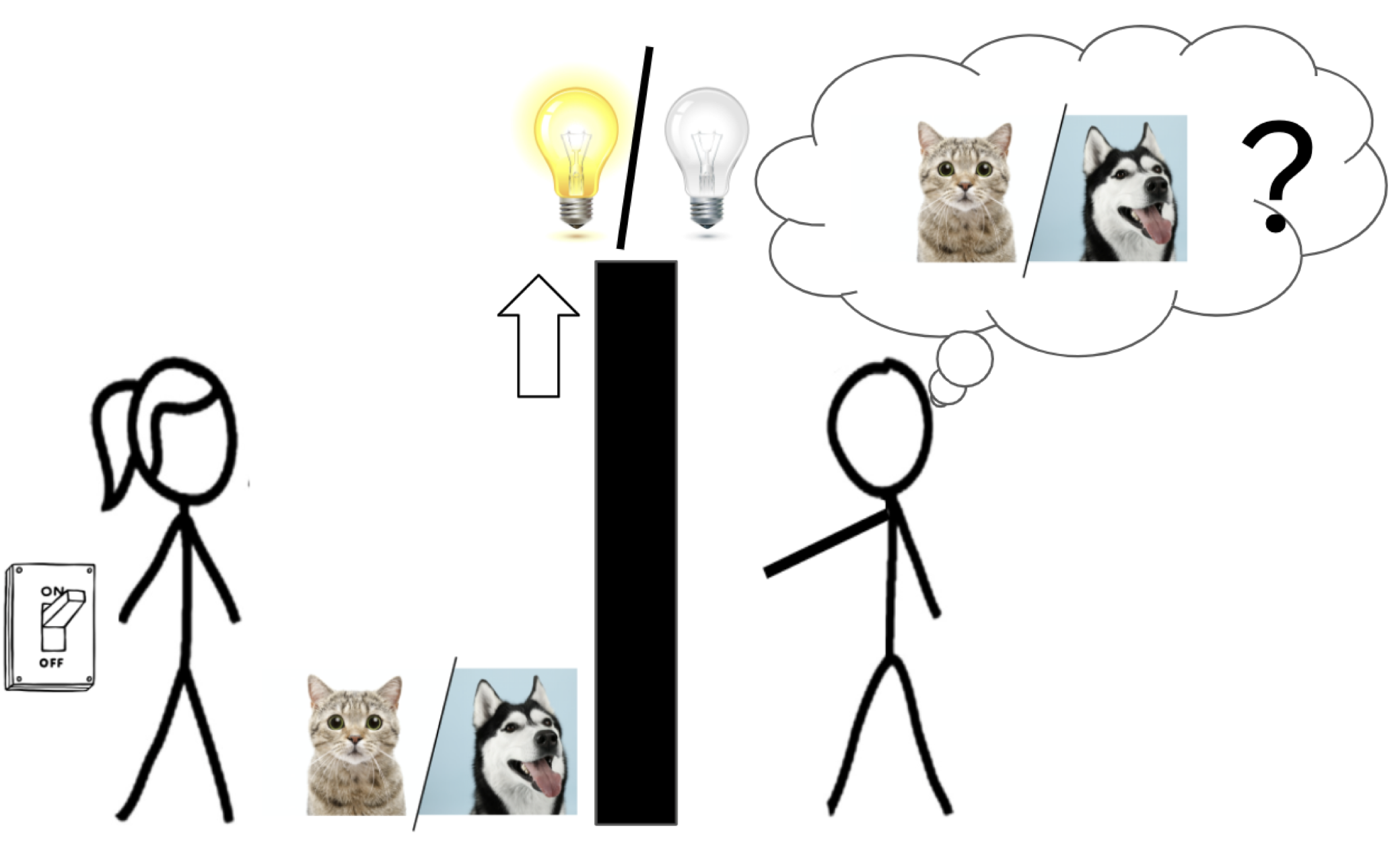}
    \includegraphics[width=0.25\textwidth]{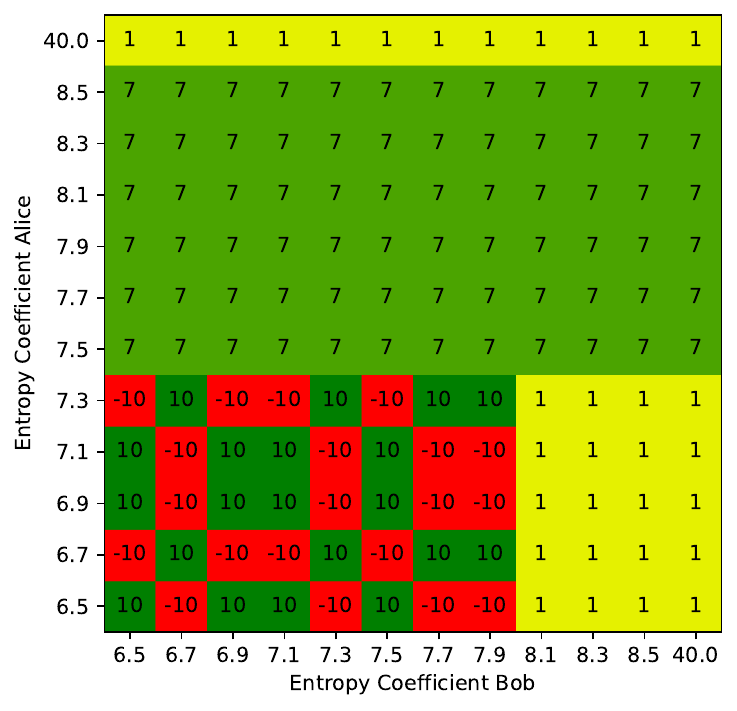}
    \includegraphics[width=0.25\textwidth]{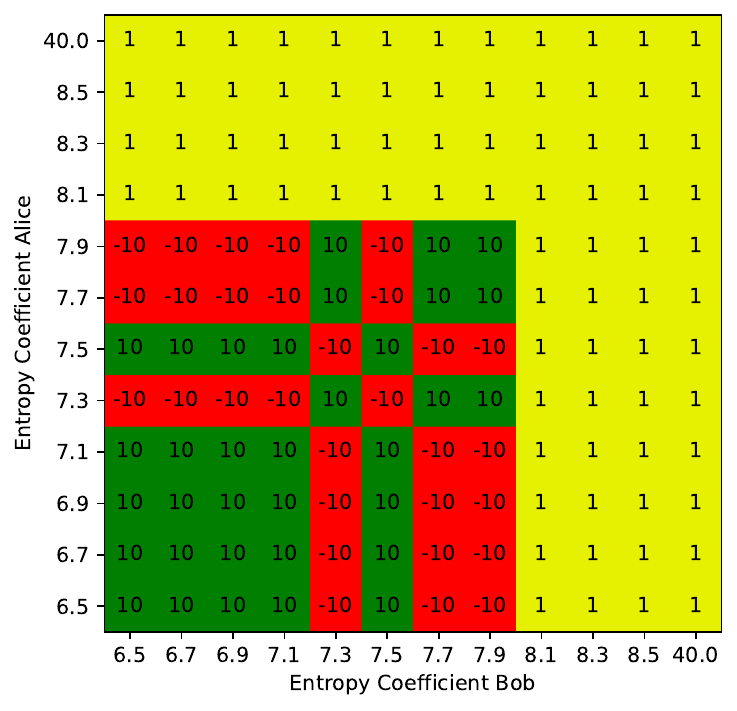}
    \caption{\textbf{Left}: Toy cooperative communication game, originally from \cite{hu2021off}. Alice observes a pet, which is either a cat or a dog with equal probability, and can then either signal ``on'', signal ``off'', ``reveal'' for a reward of $-3$, so that Bob can see the pet, or ``bail'' out for a reward of $1$. Bob then can guess ``cat'' or ``dog'' for a reward of $\pm 10$ depending on whether he was correct, or he can ``bail'' out for a reward of $1$. \textbf{Middle}: XP matrix between greedified policies in the cat/dog game, which are trained with entropy regularized independent REINFORCE with baseline, with different $\alpha$. \textbf{Right}: same as middle, except that the ``reveal'' reward is now $-8$ instead of $-3$. \textbf{Takeaway}: For sufficiently high $\alpha$ the policies stop breaking the symmetry and are fully compatible. When ``reveal'' costs $3$, one can find the optimal symmetry-equivariant policy this way, while when ``reveal'' costs $8$ one can not.}
    \label{fig:XP_cat/dog}
\end{figure}

Consider the cat/dog game from \cite{hu2021off}, described in Figure \ref{fig:XP_cat/dog}. There are two joint policies which maximize $J_{\text{SP}}$, with a return of $10$: one under which Alice always signals ``on'' when she sees a cat and ``off'' when she sees a dog, and one under which this pairing is opposite. There is one non-trivial Dec-POMDP symmetry which maps those two policies to one another, see Appendix \ref{appendix:cat/dog}, and $J_{\text{XP}}$ of these two joint policies is $-10$. Thus $J_{\text{OP}}$ of both those policies is $0$. Under the optimal symmetry-equivariant joint policy, Alice always reveals and Bob then guesses the correct pet, which means $J_{\text{SP}}$ and $J_{\text{OP}}$ equal $7$. There is also a suboptimal symmetry-equivariant policy under which Alice always bails, for a return of $1$. In the middle of Figure \ref{fig:XP_cat/dog} is shown the XP matrix between multiple greedified joint policies trained with different $\alpha$. We can deduce that for $\alpha \in \{8.1, 8.3, 8.5\}$, both Alice and Bob learn to assign the highest probability to the optimal symmetry-equivariant actions. However, we also deduce that for $\alpha \in \{7.5, 7.7, 7.9\}$, Alice assigns the highest probability to revealing, but that when Bob sees ``on'' or ``off'' he breaks the symmetry and guesses a pet instead of bailing, implying that Alice also breaks the symmetry among her suboptimal actions ``on'' and ``off''. This means that one does not have to increase $\alpha$ to the point where the non-greedy policies are symmetry-equivariant, but only to the point where the greedified policies are symmetry-equivariant.

\paragraph{Limitation: Coordination might be impossible above the entropy threshold:}

Consider a version of the cat/dog game where the reward for revealing the pet is $-8$ instead of $-3$. This gives the optimal symmetry-equivariant policy a return of $2$. The XP matrix for policies trained with different entropy coefficients is shown on the right in Figure \ref{fig:XP_cat/dog}. We see that above the entropy threshold, where the greedified policies stop breaking the symmetry, Alice already prefers bailing over revealing, since she cannot rely on Bob to choose the correct pet when she pays the high cost of revealing. Thus this example demonstrates that there exist Dec-POMDPs in which one cannot learn the optimal symmetry-equivariant policy through high entropy regularization alone: when training with low entropy one breaks the symmetry, and when training with high entropy one cannot learn the optimal symmetry-equivariant policy, since that policy requires certain actions to be chosen almost deterministically. 
For another example of a Dec-POMDP in which one cannot learn the optimal symmetry-equivariant policy through sufficiently high entropy regularization, see Appendix \ref{appendix:second toygame}.

\section{Experimental Results}

\begin{figure}
    \begin{center}
        \includegraphics[scale=0.4]{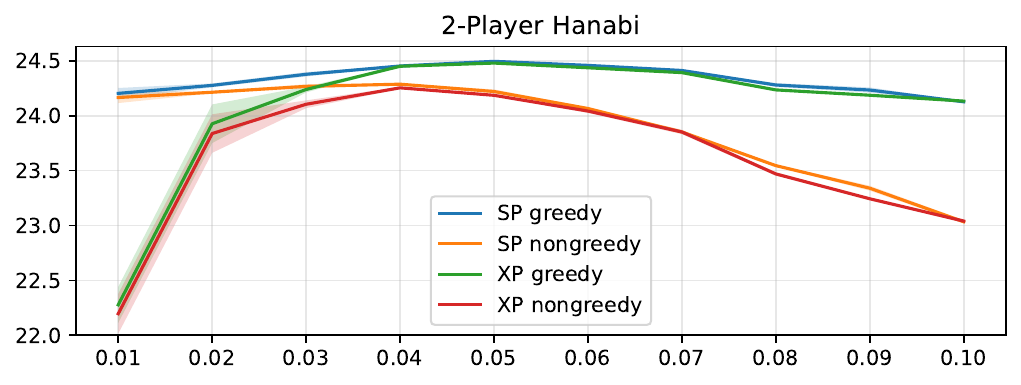}
        \hspace*{1.5mm} \includegraphics[scale=0.4]{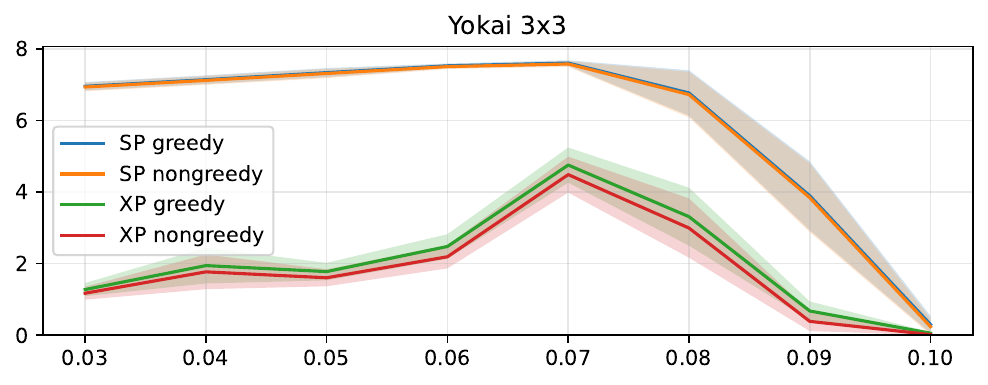}
        \includegraphics[scale=0.4]{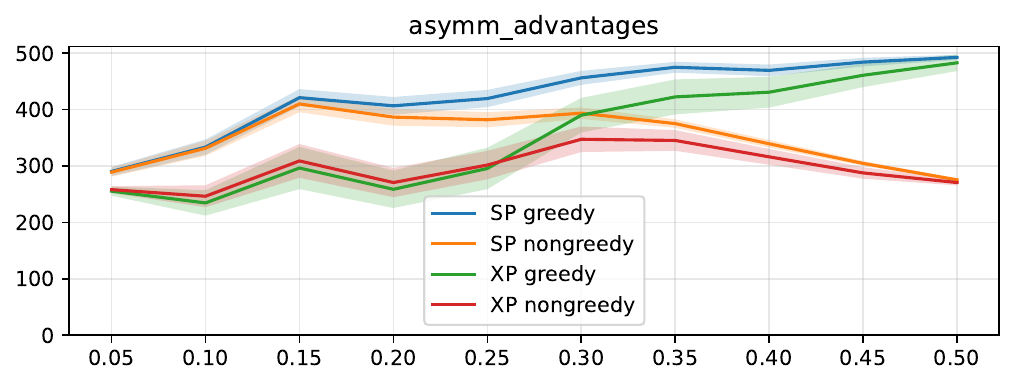}
        \includegraphics[scale=0.4]{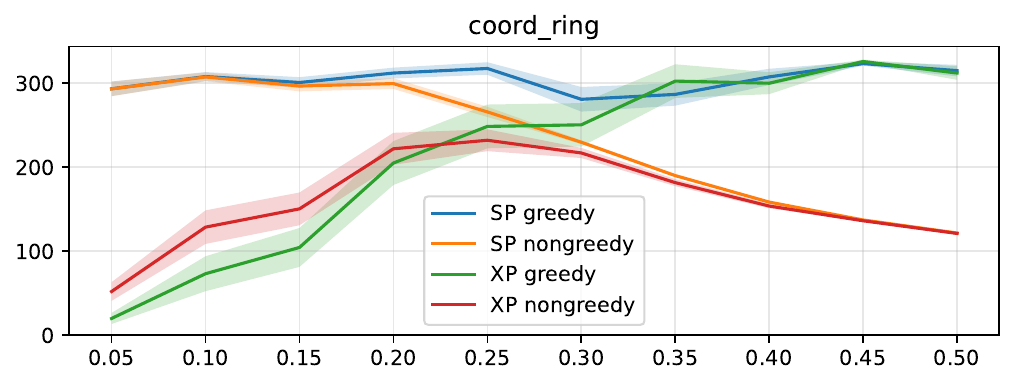}
        \includegraphics[scale=0.4]{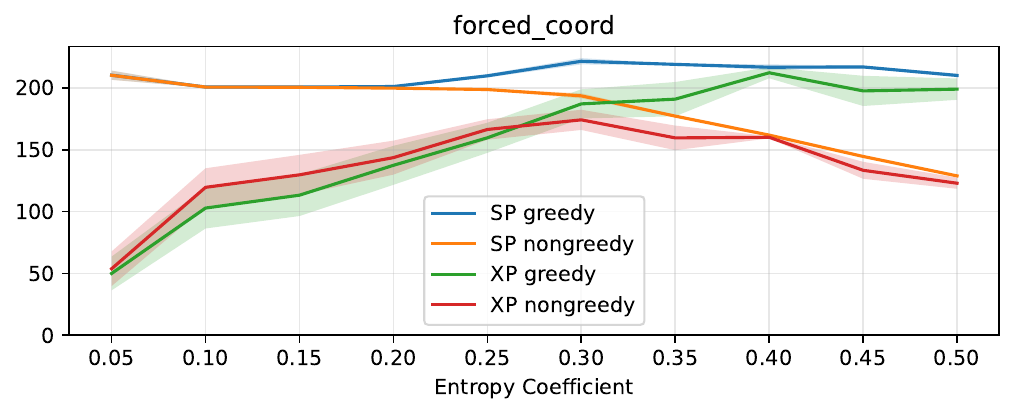}
        \includegraphics[scale=0.4]{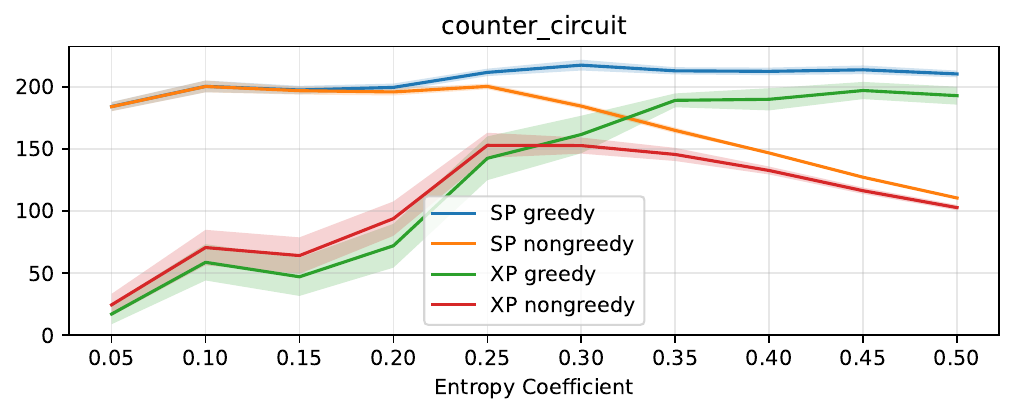}
    \end{center}
    \caption{2P Hanabi, Yokai, and four of the five standard Overcooked layouts (for ``cramped room'' see Appendix \ref{appendix:cramped room}): Mean SP and XP, greedy and non-greedy, of IPPO as a function of the entropy coefficient $\alpha$. The shaded regions show one standard error of the mean, estimated with (\ref{eq:mean XP correct}) and (\ref{eq:sigma XP correct}). We used $\lambda_{\text{GAE}} = 0.9$, $0.85$, $0.8$, and $4$, $16$, $48$ seeds per $\alpha$, for Hanabi, Yokai, Overcooked, respectively. \textbf{Takeaway}: As we increase $\alpha$, non-greedy XP approaches non-greedy SP, greedy XP approaches greedy SP, and (except in Yokai) greedy SP/XP are much higher than non-greedy SP/XP.}
    \label{fig:XP_as_function_of_entropy}
\end{figure}

\begin{figure}
    \begin{center}
        \includegraphics[scale=0.45]{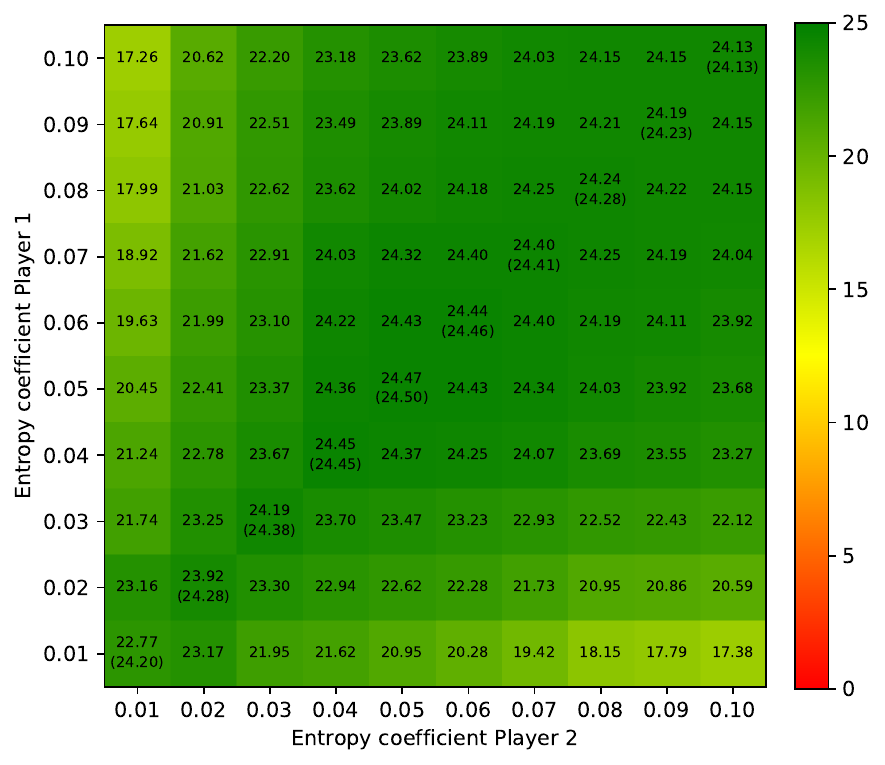}
        \includegraphics[scale=0.45]{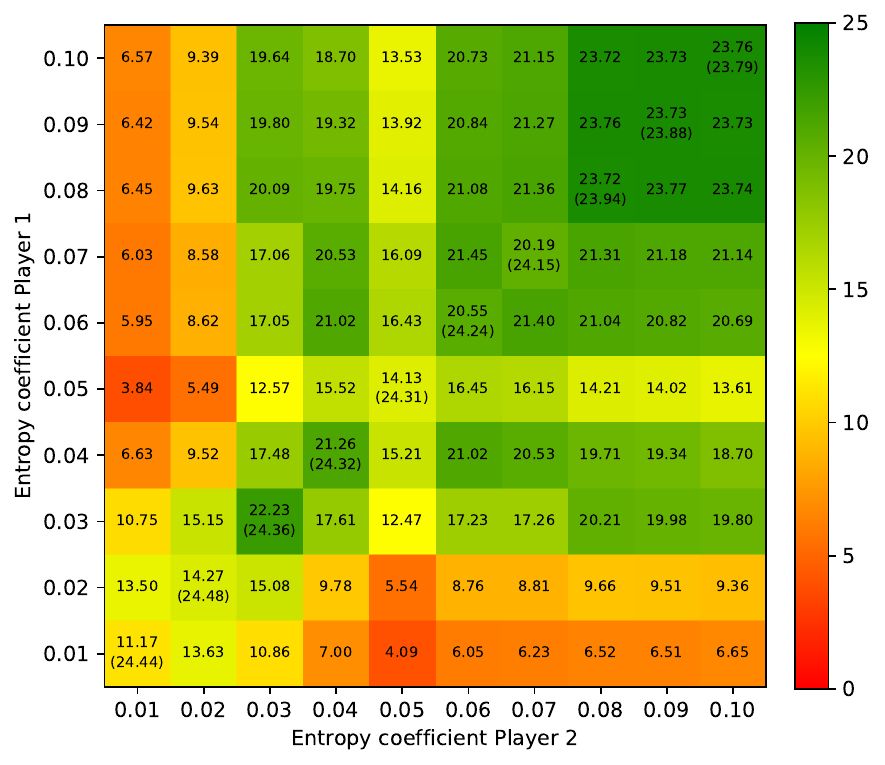}
    \end{center}
    \caption{2-Player Hanabi: Block XP matrices between 40 greedified policies trained with IPPO with FF and LSTM architectures, and different $\alpha$. Four seeds per $\alpha$. Computed from the full XP matrices in Figures \ref{fig:XP_IPPO_LSTM_FULL}, \ref{fig:XP_IPPO_FF_FULL} in Appendix \ref{appendix:full XP matrices}. On the diagonal, the numbers are the average SP (in parentheses) and the average of the off-diagonal entries in the diagonal 4x4 blocks. The off-diagonal numbers are the average of the off-diagonal 4x4 blocks. \textbf{Left}: LSTM. \textbf{Right}: FF. \textbf{Takeaway}: Policies coming from different $\alpha$'s are almost perfectly compatible when the $\alpha$'s are sufficiently high and close to each other. FF policies need much higher $\alpha$'s to be compatible than LSTM policies.}
    \label{fig:XP_IPPO}
\end{figure}

We run experiments in the JaxMARL \cite{rutherford2024jaxmarlmultiagentrlenvironments} implementations of the popular Hanabi \cite{bard2020hanabi} and Overcooked \cite{carroll2019utility} environments, as well as the recently developed Yokai environment \cite{ruhdorfer2025yokailearningenvironmenttracking}. We first investigate whether the theoretical prediction of Theorem \ref{thm:unique limit} appears in these environments, by investigating the effect of high entropy regularization during training and greedification after training, see Figure \ref{fig:XP_as_function_of_entropy}. Then we also investigate XP between policies trained with different $\alpha$, see Figure \ref{fig:XP_IPPO}. Finally we rerun seeds with the $\alpha$ that performed best in XP in Hanabi, and obtain a new SOTA in inter-seed XP. We use IPPO as it is the standard baseline deep MARL policy gradient algorithm, and its entropy bonus corresponds to our entropy regularization. The IPPO hyperparameters that are constant across all our experiments are in Appendix \ref{appendix:hyperparameters}. For every $\pi$, $J_{\text{SP}}(\pi)$ is estimated as the mean over 5000 games.

\paragraph{Finding the entropy threshold:} We first investigate the entropy thresholds of Hanabi, Overcooked and Yokai, by plotting the greedy and non-greedy SP and XP, of multiple joint policies trained with IPPO, as a function of $\alpha$. Those results are shown in Figure \ref{fig:XP_as_function_of_entropy}. We see in all environments, that as $\alpha$ increases the non-greedy SP and the non-greedy XP eventually meet, but that they decrease for high $\alpha$, exactly as Theorem \ref{thm:unique limit} guarantees in the tabular setting. We also see that greedy SP and XP eventually meet, and that, in all environments except Yokai, they are significantly higher than non-greedy SP and XP when $\alpha$ is high. In Overcooked we see that greedy XP, and sometimes even greedy SP, can keep improving for $\alpha$ going up to extremely high values of $0.50$. In Yokai the gap between SP and XP is closed only when $\alpha$ is so high that the policies cannot effectively learn anymore. While there might be other hyperparameters that achieve higher XP, our results suggest that the entropy threshold for Yokai is so high that above it no coordination is possible anymore.

We note that outside the idealized setting of Theorem \ref{thm:unique limit} the entropy threshold might depend on other hyperparameters: when using e.g. a neural network to parametrize the policy instead of the tabular parametrization (which is not possible in large Dec-POMDPs), then the entropy threshold might depend on the parametrization, as one can see in Figure \ref{fig:XP_IPPO} below. The entropy threshold can also depend on $\lambda_{\text{GAE}}$ in the generalized advantage estimates (GAE) \cite{schulman2015high}, see Appendix \ref{appendix:GAE}.

\paragraph{XP between different entropy coefficients:} For 2-player Hanabi and Overcooked we also measure XP between policies coming from different $\alpha$. For Hanabi, see Figure \ref{fig:XP_IPPO}, and for Overcooked see Appendix \ref{appendix:Overcooked Block XP}.  As in Figure \ref{fig:XP_as_function_of_entropy} we see in Figure \ref{fig:XP_IPPO} that as $\alpha$ increases greedy XP approaches greedy SP. We also see that policies coming from different $\alpha$'s are almost perfectly compatible when the $\alpha$'s are sufficiently high and close to each other. We also see that the FF policies require a much higher $\alpha$ to be compatible with each other. This can be interpreted as FF policies struggling to understand what information their partner intends to convey, since they only remember their last observation. Thus they must rely on very specialized conventions in order to communicate through this bottleneck. 

Figures \ref{fig:XP_IPPO} and \ref{fig:XP_Overcooked} were obtained in the following way: for Hanabi we use 3 different network architectures. The first architecture, referred to as ``LSTM'', consists of one feedforward embedding layer, two LSTM layers, and then one actor and one critic head. The second architecture, which we refer to as ``FF'', has separate networks for the actor and the critic, with three feedforward layers each. Third, the public-private LSTM architecture (``PP LSTM'') described in Figure \ref{fig:public-private LSTM} in Appendix \ref{appendix:hyperparameters}. We use PP LSTM in order for the actor architecture to be the same one that is used for Off-Belief Learning (OBL) \cite{hu2021off, cui2022offteamlearning}. Note that in Figure \ref{fig:XP_as_function_of_entropy} we used LSTM for Hanabi. In 2-player Hanabi, we train four seeds of IPPO policies per $\alpha$, for each of the first three architectures. For each of the three architectures we obtain $40$ joint policies $\pi_1, ..., \pi_{40}$ this way, and compute their greedy XP matrix. These XP matrices are shown in Figures \ref{fig:XP_IPPO_LSTM_FULL}, \ref{fig:XP_IPPO_LSTM_PP_FULL}, and \ref{fig:XP_IPPO_FF_FULL} in Appendix \ref{appendix:full XP matrices}. When taking the average SP and XP in the diagonal $4 \times 4$ blocks (same $\alpha$), and the average XP in the off-diagonal blocks (different $\alpha$) of those XP matrices, we get the block XP matrices shown in Figure \ref{fig:XP_IPPO}. See Appendix \ref{appendix:Overcooked Block XP} for the corresponding results for Overcooked, with 10 seeds per $\alpha$ and $10 \times 10$ blocks.

\paragraph{Hanabi: New SOTA in inter-seed XP:} The highest XP for the LSTM architecture in Figure \ref{fig:XP_IPPO} is achieved with $\alpha = 0.05$. In order to avoid selection bias, we ran new sets of 4 seeds with $\alpha = 0.05$, for IPPO with the LSTM and PP LSTM architectures. We also implemented an IPPO-based version of OBL \cite{hu2021off} and computed 4 seeds each up to level 6 and level 4 for 2- and 3-player Hanabi, respectively. For 4 and 5 players, due to computational constraints, we only show results for IPPO with the two LSTM architectures, and used $\alpha = 0.05, 0.08$ and 4, 5 seeds, respectively. We also included IPPO with LSTM architecture and $\alpha = 0.01$ to demonstrate that also for 3, 4, 5 players a low $\alpha$ leads to a large SP-XP gap. The mean SP and XP and their standard errors\footnote{We use the naive estimators (\ref{eq:mean XP naive}) and (\ref{eq:sigma XP naive}) to align with the prior work we compare against, and due to computational constraints: on an L40s, one seed of IPPO with our LSTM architecture takes more than 40, 60, 80, 100 hours, for 2, 3, 4, 5 players, respectively, and one seed of 2P OBL L6 or 3P OBL L4 takes more than 240 hours to compute.} are shown in Table \ref{table:results}. We see that for 3, 4, 5 players there are small gaps between SP and XP, indicating that the entropy thresholds (given our other hyperparameters) are slightly higher than the $\alpha$'s we used.

We see for 2 players, that XP of IPPO with either LSTM architecture is above $24.45$, which is higher than our OBL level 6 and the previous SOTA of $24.30$, achieved in \cite{cui2022offteamlearning}. Note that the highest return one can achieve in one episode of Hanabi is 25, and one cannot achieve that return in every episode. Thus the maximum possible expected return of any joint policy is lower than 25. Given our low standard errors, the improvement from $24.30$ to $24.45$ appears statistically significant. Also for 3 players we see that XP of IPPO with $\alpha = 0.05$ and PP LSTM architecture is higher than our OBL level 4 and the previous SOTA of inter-seed XP in 3 player Hanabi of $23.02$, achieved in \cite{hu2021off}.

\begin{table}[ht]
\centering
\caption{Hanabi: Mean $\pm$ standard error of greedy SP and XP, estimated with (\ref{eq:mean XP naive}) and (\ref{eq:sigma XP naive}), of IPPO with LSTM and PP LSTM architectures, with different $\alpha$. Also included are: our own implementation of OBL, OT-OBL L5 from \cite{cui2022offteamlearning} (where only XP is reported), and OBL L4 from the original OBL paper \cite{hu2021off}. For 2P, 3P, 4P we computed 4 seeds with each algorithm, and for 5P we computed 5 seeds. \textbf{Takeaway}: IPPO with $\alpha=0.01$ has the highest SP, but very low XP, while for $\alpha = 0.05$ ($0.08$ for 5P) SP is slightly lower than for $0.01$, but XP is almost as high as SP, and for 2P higher than XP of all 3 versions of OBL, the previous SOTA in inter-seed XP in 2P Hanabi.}
\label{table:results}
\begin{tabular}{l|cc|cc}
\hline
 & 2P SP & 2P XP & 3P SP & 3P XP \\
\hline
IPPO $\alpha=0.01$ & $\textbf{24.51} \pm 0.02$ & $15.45 \pm 1.89$ & $\textbf{24.76} \pm 0.02$ & $3.57 \pm 0.4$ \\
IPPO $\alpha=0.05$ & $24.48 \pm 0.02$ & $24.45 \pm 0.02$ & $24.48 \pm 0.02$ & $24.28 \pm 0.03$ \\
IPPO PP $\alpha=0.05$ & $24.49 \pm 0.02$ & $\textbf{24.46} \pm 0.02$ & $24.66 \pm 0.02$ & $\textbf{24.55} \pm 0.03$ \\
OBL (ours), L6 (2P) / L4 (3P)  & $24.35 \pm 0.02$ & $24.30 \pm 0.02$ & $24.48 \pm 0.01$ & $24.47 \pm 0.02$ \\
OBL (original) L4  & $24.10 \pm 0.01$ & $23.76 \pm 0.06$ & $23.38 \pm 0.04$ & $23.02 \pm 0.01$ \\
OT-OBL L5 & - & $24.30 \pm 0.01$ & - & - \\
\hline
 & 4P SP & 4P XP & 5P SP & 5P XP \\
\hline
IPPO $\alpha=0.01$ & $\textbf{24.59} \pm 0.04$ & $0.31 \pm 0.02$ & $\textbf{24.09} \pm 0.06$ &  $0.32 \pm 0.01$ \\
IPPO $\alpha=0.05$ / $\alpha=0.08$ & $24.34 \pm 0.09$ & $24.10 \pm 0.03$ & $23.32 \pm 0.35$ & $21.27 \pm 0.12$ \\
IPPO PP $\alpha=0.05$ / $\alpha=0.08$ & $24.55 \pm 0.01$ & $\textbf{24.30} \pm 0.03$ & $23.73 \pm 0.02$ & $\textbf{23.59} \pm 0.03$ \\
\hline
\end{tabular}
\end{table}

\section{Related Work}

\cite{agarwal2020, mei2020, ding2025} studied convergence rates of tabular softmax policy gradient ascent to an optimal policy in single-agent MDPs, and \cite{zhang2022} studies the (rate of) convergence towards a Nash equilibrium in fully observable potential games (which include fully observable Dec-POMDPs). Those papers, and also \cite{zafarali2019}, also study the effect of various forms of entropy regularization on the convergence rate, but only forms which are gradients of an objective function, e.g. the maximum entropy RL objective. \cite{balduzzi2018} studies gradient-based methods in general-sum games, but just as \cite{zhang2022} is not concerned \textit{which} Nash equilibrium one converges to. In \cite{rudolph2025reevaluatingpolicygradientmethods} it was found that in various imperfect information two-player zero-sum games, generic IPPO with increased entropy regularization beats, in terms of exploitability of the trained policies, other popular algorithms which were specifically designed for this setting. In \cite{lauffer2025robustdiversemultiagentlearning}, a new MARL policy gradient method called \textit{rational policy gradient} is developed for Dec-POMDPs, which aims to learn policies that are robust and diverse. They also briefly mention that in their experiments in the reduced 3- and 4-color versions of 2-player Hanabi, independent policy gradient ascent with $\alpha = 0.05$ led to higher inter-seed XP than with $\alpha = 0.01$. They speculate that this is a feature of the smaller game or due to the fact that they use FF policies without history dependence, but do not expand on it further. 

\section{Conclusion}

Both our theoretical and empirical results show that standard policy gradient methods in Dec-POMDPs are much more powerful for XP than previously assumed: with high entropy regularization during and greedification after training, they can be very effective in learning high-performing symmetry-equivariant policies, and this recipe can have a massive influence on inter-seed XP. Despite examples of Dec-POMDPs in which one can not learn the optimal symmetry-equivariant policy this way, our results suggest that during hyperparameter sweeps in Dec-POMDPs one should not only evaluate SP but also XP, and consider a much bigger range of entropy coefficients than is typically done.

\newpage

\subsubsection*{Acknowledgments}
J. Foerster is partially funded and J. Forkel is fully funded by the UKRI grant EP/Y028481/1 (originally selected for funding by the ERC). 
The authors thank the International Max Planck Research School for Intelligent Systems (IMPRS-IS) for supporting C. Ruhdorfer. M. Beukman is supported by the Rhodes Trust.

\section*{Errata}

In a previous arXiv version our Theorem \ref{thm:unique limit} also stated that above the entropy threshold, policy gradient ascent with non-infinitesimal steps, i.e. the sequence (\ref{eq:policy gradient ascent}), with unbiased but finite-variance estimates of the gradients, will almost surely converge to $\theta^\alpha$ for any $\theta_0 \in \Theta$. While we still conjecture this to be always true, there was a mistake in that part of our proof, specifically in the construction of a Lyapunov function which is needed to apply the stochastic approximation results from \cite{borkar2023}. We did not manage to repair that mistake (yet), so now our Theorem \ref{thm:unique limit} only guarantees convergence when following the exact policy gradient ODE.

\bibliography{main}
\bibliographystyle{plainnat}

\newpage
\appendix
\onecolumn

\section{Proof of Theorem \ref{thm:unique limit}} \label{appendix:proof}

In order to prove Theorem \ref{thm:unique limit} we state and prove the following two auxiliary lemmas.

\begin{lemma} \label{lemma:bounded logits}
    Let $\pi_\theta = \text{softmax}(\theta) \in (0, 1)^d$ for $\theta \in \mathbb{R}^d$, let $\alpha > 0$, let $x \mapsto r_x \in [-1, 1]^d$ be a continuous function, and let $x \mapsto \theta_x$ be the solution to the ODE
    \begin{align}
        \frac{\text{d} \theta_x}{\text{d}x} = \nabla_\theta \left( \pi_{\theta_x}^T (r_x - \alpha \log \pi_{\theta_x} ) \right), \, \, x \geq 0,
    \end{align}
    for a starting point $\theta_0$ which satisfies $\sum_{a = 1}^d \theta_0(a) = 0$. Then $\sum_{a = 1}^d \theta_x(a) = 0$ for all $x \geq 0$, and there exists $x' \in [0, \infty)$, such that $x \geq x' \implies \| \theta_x \|_\infty < 2d \alpha^{-1}$.
\end{lemma}

\begin{proof}
    We see that    
    \begin{align}
    \begin{split}
        \frac{\text{d}\theta_x}{\text{d}x} =& \pi_{\theta_x} \odot \left( r_x - \alpha \log \pi_{\theta_x} - \mathbb{E}_{\pi_{\theta_x}} \left[ r_x \right] - \alpha \text{Ent}(\pi_{\theta_x}) \right) \\
        =& \pi_{\theta_x} \odot \left( r_x - \alpha \theta_x - \mathbb{E}_{\pi_{\theta_x}} \left[ r_x - \alpha \theta_x \right] \right).
    \end{split}
    \end{align}
    Thus $\frac{\text{d}}{\text{d}x} \sum_{a = 1}^d \theta_x(a) = \sum_{a = 1}^d \frac{\text{d}}{\text{d}x} \theta_x(a) = \mathbb{E}_{\pi_{\theta_x}} \left[ r_x - \alpha \theta_x \right] - \mathbb{E}_{\pi_{\theta_x}} \left[ r_x - \alpha \theta_x \right] = 0$ for all $x \geq 0$, which together with the assumption that $\sum_{a = 1}^d \theta_0(a) = 0$ implies that $\sum_{a = 1}^d \theta_x(a) = 0$ for all $x \geq 0$. Furthermore, we see that $\mathbb{E}_{\pi_{\theta}}\left[ \theta \right] \geq \frac{1}{d} \sum_{a = 1}^d \theta(a)$ for all $\theta \in \mathbb{R}^d$, since for $g(\beta) := \langle \pi_{\beta \theta}, \theta \rangle$ and $H(\pi_\theta) := \frac{\text{d} \pi_\theta}{\text{d}\theta} = \text{Diag}(\pi_\theta) - \pi_\theta \pi_\theta^T \in \mathbb{R}^{d \times d}$ it holds that
    \begin{align}
        g(0) = \frac{1}{d} \sum_{a = 1}^d \theta(a), \quad g(1) = \mathbb{E}_{\pi_{\theta}} \left[ \theta \right], \quad \frac{\text{d}g}{\text{d}\beta} (\beta) = \langle \theta, H(\pi_{\beta\theta}) \theta \rangle = \text{Var}_{\pi_{\beta\theta}} \left[ \theta \right] \geq 0.
    \end{align}
    Therefore, since $r_x \in [-1, 1]^d$, we see that
    \begin{align}
        r_x(a) - \alpha \theta_x(a) - \mathbb{E}_{\pi_{\theta_x}} \left[ r_x - \alpha \theta_x \right] \geq - 2 - \alpha \theta_x(a), \, \, a = 1, ..., d.
    \end{align}
    Thus for any $\epsilon > 0$ we see that
    \begin{align}
        \theta_x(a) \leq -(2 + \epsilon) \alpha^{-1} \implies \frac{\text{d}\theta_x}{\text{d}x}(a) \geq \epsilon \pi_{\theta_x}(a) > 0,
    \end{align}
    which implies that for all $\theta_0$ there exists an $x' \geq 0$ such that $\min_{a = 1, ..., d} \theta_x(a) \geq - (2 + \epsilon) \alpha^{-1}$ for all $x \geq x'$. Furthermore, since $\sum_{a = 1}^d \theta_x(a) = 0$ for all $x \geq 0$, we then see that also $\| \theta_x \|_\infty < 2d \alpha^{-1}$ for all $x \geq x'$.
\end{proof}

\begin{lemma} \label{lemma:equivariance of gradient}
    If there exists a $\theta^\alpha \in \Theta$ such that $\{ \theta \in \Theta : \nabla_\theta^\alpha J_{\text{SP}}(\pi_\theta) = 0 \} = \{ \theta^\alpha \}$, then $\phi(\pi_{\theta^\alpha}) = \pi_{\theta^\alpha}$ for all $\phi \in \Phi$.
\end{lemma}

\begin{proof}
Given centered logits $\theta \in \Theta$, a policy $\pi_\theta$ is an element of $[0, 1]^d$. Therefore, applying $\phi$ to a policy $\pi_\theta$, merely corresponds to a permutation of the probabilities, which means we can interpret $\phi$ as a $d \times d$ permutation matrix. Due to the permutation equivariance of the softmax function we can also apply $\phi$ to the logits $\theta \in \Theta$ and can write $\phi(\pi_\theta) = \pi_{\phi(\theta)}$. 

For simplicity of notation we define $F(\theta) := \nabla_\theta^\alpha J_{\text{SP}}(\pi_\theta)$. If $\phi(F(\theta)) = F(\phi(\theta))$ for some $\phi \in \Phi$ and all $\theta \in \Theta$, then
\begin{align}
    0 = F(\theta) \implies \phi(0) = \phi(F(\theta)) \implies 0 = F(\phi(\theta)) \implies \phi(\theta^\alpha) = \theta^\alpha,
\end{align}
where in the last step we have used the fact that $\theta^\alpha$ is the only zero of $F(\theta)$ in $\Theta$. We now show that $\phi(F(\theta)) = F(\phi(\theta))$ for all $\phi \in \Phi$ and all $\theta \in \Theta$.

From the definitions of Dec-POMDP symmetries and their action on policies we see that
\begin{align}
    p_{\pi_{\phi(\theta)}}(\phi(\tau_T)) =& b_0(\phi(s_0)) \prod_{t = 0}^{T-1} \phi(\pi_\theta)(\phi(a_t) | \phi(\tau_t)) \mathcal{T}(\phi(s_{t+1}) | \phi(s_t), \phi(a_t)) \\
    =& b_0(s_0) \prod_{t = 0}^{T-1} \pi_\theta(a_t | \tau_t) \mathcal{T}(s_{t+1} | s_t, a_t) = p_{\pi_\theta}(\tau_T).
\end{align}
For simplicity we introduce the following notation: $f(\theta) := J_{\text{SP}}(\pi_\theta)$, and for every SAH $\tau_t$ we define $f_{\tau_t}(\theta) = \text{Ent}(\pi_\theta( \cdot | \tau_t))$. Then we see that
\begin{align}
\begin{split}
    f(\phi(\theta)) =& J_{\text{SP}}(\phi(\pi_\theta)) \\
    =& \sum_{\tau_T} p_{\phi(\pi_\theta)}(\tau_T) \sum_{t = 0}^{T-1} \gamma^t \mathcal{R}(s_{t+1}, s_t, a_t) \\
    =& \sum_{\tau_T} p_{\phi(\pi_\theta)}(\phi(\tau_T)) \sum_{t = 0}^{T-1} \gamma^t \mathcal{R}(\phi(s_{t+1}), \phi(s_t), \phi(a_t)) \\
    =& \sum_{\tau_T} p_{\pi_\theta}(\tau_T) \sum_{t = 0}^{T-1} \gamma^t \mathcal{R}(s_{t+1}, s_t, a_t) \\
    =& J_{\text{SP}}(\pi_\theta) = f(\theta), \\
    f_{\tau_t}(\theta) =& - \sum_{a_t} \pi_\theta(a_t | \tau_t) \log \pi_{\theta}(a_t | \tau_t) \\
    =& - \sum_{a_t} \pi_{\phi(\theta)}( \phi(a_t) | \phi(\tau_t)) \log \pi_{\phi(\theta)}( \phi(a_t) | \phi(\tau_t)) = f_{\phi(\tau_t)}(\phi(\theta)).
\end{split}
\end{align}
Since $\phi$ corresponds to a permutation matrix in logit space, we see that its transpose $\phi^T$ equals its inverse $\phi^{-1}$, and that the derivative of $\theta \mapsto \phi(\theta)$ is given by $\phi$. Thus for any $\phi \in \Phi$ the chain rule implies that
\begin{align}
\begin{split}
    \nabla_\theta f(\theta) =& \nabla_\theta (f \circ \phi)(\theta) = \phi^T(\nabla_\theta f)(\phi(\theta)) = \phi^{-1}(\nabla_\theta f)(\phi(\theta)), \\
    (\nabla_\theta f_{\tau_t})(\theta) =& \phi^T (\nabla_\theta f_{\phi(\tau_t)}) (\phi(\theta)) = \phi^{-1} (\nabla_\theta f_{\phi(\tau_t)}) (\phi(\theta)).
\end{split}
\end{align}
Combining the above equations, we see that
\begin{align}
\begin{split}
    F(\theta) =& \nabla_\theta f(\theta) + \alpha \sum_{\tau_T} p_{\pi_\theta}(\tau_T) \sum_{t = 0}^{T-1} \gamma^t \nabla_\theta f_{\tau_t}(\theta) \\
    =& \phi^{-1} \nabla_\theta f(\phi(\theta)) + \alpha \sum_{\tau_T} p_{\pi_{\phi(\theta)}}(\phi(\tau_T)) \sum_{t = 0}^{T-1} \gamma^t \phi^{-1} (\nabla_\theta f_{\phi(\tau_t)}) (\phi(\theta)) \\
    =& \phi^{-1} \nabla_\theta f(\phi(\theta)) + \alpha \phi^{-1} \sum_{\tau_T} p_{\pi_{\phi(\theta)}}(\tau_T) \sum_{t = 0}^{T-1} \gamma^t (\nabla_\theta f_{\tau_t}) (\phi(\theta)) \\
    =& \phi^{-1} F(\phi(\theta)),
\end{split}
\end{align}
which finishes the proof.
\end{proof}

We now prove Theorem \ref{thm:unique limit}, which we state here again for completion.

\begin{theorem}
    Let $\theta \mapsto \pi_\theta$ be the tabular softmax parametrization in a Dec-POMDP. Then there exists a finite entropy threshold $\alpha' \in [0, \infty)$, such that for all $\alpha > \alpha'$ there exists a unique $\theta^\alpha \in \Theta$ at which the vector field $\theta \mapsto \nabla_\theta^\alpha J_{\text{SP}}(\pi_{\theta})$ equals zero. Furthermore, $\pi_{\theta^\alpha}$ is symmetry-equivariant, i.e. $\phi(\pi_{\theta^\alpha}) = \pi_{\theta^\alpha}$ for all $\phi \in \Phi$. Finally, for any $\theta_0 \in \Theta$ it holds that $\lim_{x \rightarrow \infty} \theta_x = \theta^\alpha$, where $x \mapsto \theta_x$ is the solution to the ODE $\frac{\text{d}\theta_x}{\text{d}x} = \nabla_\theta^\alpha J_{\text{SP}}(\pi_{\theta_x})$ starting at $\theta_0$.
\end{theorem}

\begin{proof}
Throughout this entire proof we consider all logits to be centered, i.e. lie in $\Theta$. We see that
\begin{align}
    & \nabla_\theta^\alpha J_{\text{SP}}(\pi_\theta) \\
    =& \nabla_\theta J_{\text{SP}}(\pi_\theta) + \alpha \mathbb{E}_{\tau_T \sim \pi_\theta} \left[ \nabla_\theta \sum_{t = 0}^{T-1} \gamma^t \text{Ent}(\pi_\theta( \cdot | \tau_t)) \right] \\
    =& \mathbb{E}_{\tau_T \sim \pi_\theta} \left[ \sum_{t = 0}^{T-1} \gamma^t \sum_{i = 1}^n \left( \alpha \nabla_\theta \text{Ent}(\pi_\theta^i(\cdot | \tau_t^i)) + \nabla_\theta \log \pi_\theta^i(a_{t}^i | \tau_{t}^i) \sum_{t' = t}^{T-1} \gamma^{t' - t} \mathcal{R}(s_{t'+1}, s_{t'}, a_{t'}) \right) \right] \\ 
    =& \sum_{t = 0}^{T-1} \gamma^t \sum_{i = 1}^n \sum_{\tau_t^i} p_{\pi_\theta}(\tau_t^i) \left( \alpha \nabla_\theta \text{Ent}(\pi_\theta^i(\cdot | \tau_t^i)) + \sum_{a_t^i \in \mathcal{A}^i(o_t^i)} \nabla_\theta \pi_\theta^i(a_t^i | \tau_t^i) Q_{\pi_\theta}^i(a_t^i | \tau_t^i) \right) \\
    =& \sum_{t = 0}^{T-1} \gamma^t \sum_{i = 1}^n \sum_{\tau_t^i} p_{\pi_\theta}(\tau_t^i) \sum_{a_t^i \in \mathcal{A}^i(o_t^i)} \nabla_\theta \pi_\theta^i(a_t^i | \tau_t^i) \left( - \alpha \theta^i(a_t^i | \tau_t^i)  + Q_{\pi_\theta}^i(a_t^i | \tau_t^i) \right), \label{eq:Q values}
\end{align}
where $Q_{\pi_\theta}^i(a_t^i | \tau_t^i) = \mathbb{E}_{\tau_T \sim \pi_\theta} \left[ \sum_{t' = t}^{T-1} \gamma^{t' - t} \mathcal{R}(s_{t'+1}, s_{t'}, a_{t'}) | a_t^i, \tau_t^i \right]$. Define the map
\begin{align}
    G: [0, \infty) \times \Theta &\rightarrow \Theta = \left\{ \theta \in \mathbb{R}^d: \forall i, \tau^i: \sum_{a^i \in \mathcal{A}^i(o_t^i)} \theta^i(a^i | \tau^i) = 0 \right\} \\
    (\epsilon, \theta) &\mapsto \epsilon \nabla_\theta J_{\text{SP}}(\pi_\theta) + \mathbb{E}_{\tau_T \sim \pi_\theta} \left[ \nabla_\theta \sum_{t = 0}^{T-1} \gamma^t \text{Ent}(\pi_\theta( \cdot | \tau_t)) \right],
\end{align}
which, as a composition of smooth functions, is smooth. We see that $G(0, \theta) = 0 \iff \theta = 0$ because for every $\theta \in \Theta$ every realizable local AOH $\tau_t^i$ happens with positive probability and only for $\theta = 0$ is $\nabla_\theta \text{Ent}(\pi_\theta^i(\cdot | \tau^i_t)) = 0$ for all local AOHs $\tau_t^i$. We now show that the Jacobian w.r.t. $\theta$ of $G$ is negative definite for $\epsilon = 0$ and $\theta = 0$, and thus invertible (both statements are to be understood in $\Theta$, the space of centered logits):
\begin{align}
    \text{Jac}_\theta G(0, \theta) = \sum_{t = 0}^{T-1} \gamma^t \sum_{i = 1}^n \sum_{\tau^i_t} \left( p_{\pi_\theta}(\tau^i_t) \text{Hess}_\theta \text{Ent}(\pi_\theta^i(\cdot | \tau^i_t)) + \nabla_\theta \text{Ent}(\pi_\theta^i(\cdot | \tau^i_t)) \left( \nabla_\theta p_{\pi_\theta}(\tau^i_t) \right)^T \right).
\end{align}
Since for every $\theta \in \Theta$ it holds that $p_{\pi_\theta}(\tau_t^i) > 0$ for all $\tau_t^i$, and since for $\theta = 0$ it holds that $\nabla_\theta \text{Ent}(\pi_\theta^i(\cdot | \tau^i_t)) = 0$ for all $\tau_t^i$, we see that $\text{Jac}_\theta G(0, 0)$ is negative definite as it is a block-diagonal matrix with negative definite blocks.
    
Thus by the implicit function theorem, there exist $\epsilon' > 0$ and $ R > 0$ such that there exists a unique function $g: [0, \epsilon') \rightarrow \{\theta \in \Theta: \, \|\theta \|_\infty < R \}$ for which $g(0) = 0$ and $G(\epsilon, g(\epsilon)) = 0$ for all $\epsilon \in [0, \epsilon')$. Since $\text{Jac}_\theta G(0, 0)$ is negative definite, we can assume w.l.o.g. that $\epsilon'$ and $R$ were chosen small enough such that for all $\epsilon \in [0, \epsilon')$ and $\| \theta \|_\infty < R$ it holds that $\langle \theta - g(\epsilon), G(\epsilon, \theta) \rangle < 0$.

We see that for $\alpha > 0$ it holds that $\nabla_\theta^\alpha J_{\text{SP}}(\pi_\theta) = \alpha G(\alpha^{-1}, \theta)$. We choose $\alpha' = (\epsilon')^{-1}$, and for $\alpha > \alpha'$ set $\theta^\alpha = g(\alpha^{-1})$. Then, given that $g$ is unique, we see that when $\alpha > \alpha'$ and $\| \theta \|_\infty < R$, then $\nabla_\theta^{\alpha} J_{\text{SP}}(\pi_\theta) = 0 \iff \theta = \theta^\alpha$.

We then see that for $\alpha > \alpha'$ and $\| \theta_x \|_\infty < R$ the Lyapunov function $V_\alpha(\theta) := \frac{1}{2} \| \theta - \theta^\alpha \|_2^2$ satisfies
\begin{align}
    \frac{\text{d}}{\text{d}x} V_\alpha(\theta_x) = \langle \theta_x - \theta^\alpha, \frac{\text{d}\theta_x}{\text{d}x} \rangle = \langle \theta_x - \theta^\alpha, \nabla_\theta^\alpha J_{\text{SP}}(\pi_{\theta_x}) \rangle = \alpha \langle \theta_x - \theta^\alpha, G(\alpha^{-1}, \theta_x) \rangle < 0.
\end{align}
Thus we see that if $\| \theta_x \|_\infty < R$, then $\lim_{x \rightarrow \infty} V_\alpha(\theta_x) = 0$ and thus $\lim_{x \rightarrow \infty} \theta_x = \theta^\alpha$.

We now further assume w.l.o.g. that $\alpha'$ was chosen such that
\begin{align}
     \alpha' \geq 2 R^{-1} \max_{\tau_t^i, a_t^i \in \mathcal{A}^i(o_t^i), \, \theta \in \Theta} |\mathcal{A}^i(o_t^i)| \cdot \left| Q_{\pi_\theta}^i(a_t^i | \tau_t^i) \right|,
\end{align}
where  the lower bound on the RHS is finite since the state, action and observation spaces, the rewards, and the time horizon are all finite. Applying Lemma \ref{lemma:bounded logits} to every $\tau_t^i$, iteratively for $t = 0, ..., T-1$, we see that for any $\alpha > \alpha'$ and $\theta_0$ there exists $x' \geq 0$, such that for all $x \geq x'$ it holds that $\| \theta_x \|_\infty < R$. This in particular implies that for $\alpha > \alpha'$ and $\| \theta \|_\infty \geq R$ it holds that $\nabla_\theta^\alpha J_{\text{SP}}(\pi_\theta) \neq 0$. Since we have already established that $\lim_{x \rightarrow \infty} \theta_x = \theta^\alpha$ if $\| \theta_x \|_\infty < R$ for some $x \geq 0$, we see that for any $\theta_0$ it holds that $\lim_{x \rightarrow \infty} \theta_x = \theta^\alpha$. Therefore $\{ \theta \in \Theta : \nabla_\theta^\alpha J_{\text{SP}}(\pi_\theta) = 0 \} = \{ \theta^\alpha \}$, which by Lemma \ref{lemma:equivariance of gradient} implies that $\phi(\pi_{\theta^\alpha}) = \pi_{\theta^\alpha}$ for all $\phi \in \Phi$. This finishes the proof.
\end{proof}

\newpage

\section{How to estimate the mean and standard deviation of $J_{\text{XP}}$} \label{appendix:XP estimators}

In this section we treat the noise in the estimation of $J_{\text{SP}}(\pi)$ for a computed joint policy $\pi$ as negligible. This is because in practice $J_{\text{SP}}(\pi)$ is evaluated as the average return over a number of episodes, which is very computationally cheap. E.g. computing the average return over 5000 episodes in Hanabi takes about 0.25 seconds with our code. Therefore we assume that we can make that noise negligibly small.

Given an $n$-player Dec-POMDP and algorithms $L_1, ..., L_n$, which we view as random variables with values in $\Pi$, we get the random variable $J_{\text{XP}}(L_1, ..., L_n)$. ZSC corresponds to the setting where $L_1, ..., L_n \stackrel{\mathrm{iid}}{\sim} L$ are independent and identically distributed, i.e. $L_1, ..., L_n$ are independent copies of some algorithm $L$. Given such an algorithm $L$, we are interested in obtaining estimates of the mean, denoted $\mu$, and the standard deviation, denoted $\sigma$, of $J_{\text{XP}}(L_1, ..., L_n)$. 

The most natural way to obtain such estimates, which is commonly used \cite{hu2020other, cui2021k, hu2021off, cui2022offteamlearning, lupu2021trajectory, muglich2022equivariant, muglich2025expectedreturnsymmetries, lauffer2025robustdiversemultiagentlearning}, but statistically not entirely correct, is to run the algorithm $m \geq n$ times to produce joint policies $\pi_1, ..., \pi_m$, and then estimate $\mu$ and $\sigma^2$ as
\begin{align}
    \hat{\mu}_{\text{naive}} :=& \frac{(m-n)!}{m!} \sum_{j \in I(m, n)} J_{\text{SP}}((\pi_{j_1}^1, ..., \pi_{j_n}^n)), \label{eq:mean XP naive} \\
    \hat{\sigma}_{\text{naive}}^2 :=& \frac{(m-n)!}{m! - (m-n)!} \sum_{j \in I(m, n)} \left( J_{\text{SP}}((\pi_{j_1}^1, ..., \pi_{j_n}^n)) - \hat{\mu}_{\text{naive}} \right)^2, \label{eq:sigma XP naive}
\end{align}
where $I(m, n) := \{j \in \{1, ..., m\}^n:  j_k \neq j_l \text{ if } k \neq l\}$. In the case $n=2$, this just corresponds to taking the mean and the Bessel-corrected sample variance of the off-diagonal entries in the XP matrix $\left( J_{\text{SP}}((\pi_j^1, \pi_k^2)) \right)_{j,k = 1}^{m}$.

While $\hat{\mu}_{\text{naive}}$ is an unbiased estimator of $\mu$, the problem here is that $\hat{\sigma}_{\text{naive}}^2$ is in general \textbf{not} an unbiased estimator of $\sigma^2$, since the summands in (\ref{eq:mean XP naive}) are not sampled independently. Thus $\frac{(m-n)!}{m!} \hat{\sigma}^2_{\text{naive}}$ is \textbf{not} an unbiased estimate of the squared standard error of $\hat{\mu}_{\text{naive}}$.

We propose the following unbiased estimators of $\mu$ and $\sigma^2$: in order to get $m$ independent samples of $J_{\text{XP}}(L_1, ..., L_n)$, we need to compute $m \times n$ joint policies $\pi_{i, j}$, $i=1, ..., n$, $j = 1,..., m$. Then $J_{\text{XP}}(\pi_{1, j}, ..., \pi_{n, j})$, $j = 1,..., m$, are independent samples from which we can compute the usual unbiased estimators:
\begin{align}
    \hat{\mu} :=& \frac{1}{m} \sum_{j = 1}^m J_{\text{XP}}(\pi_{1, j}, ..., \pi_{n, j}), \label{eq:mean XP correct} \\
    \hat{\sigma}^2 :=& \frac{1}{m-1} \sum_{j = 1}^m \left( J_{\text{XP}}(\pi_{1, j}, ..., \pi_{n, j}) - \hat{\mu} \right)^2. \label{eq:sigma XP correct}
\end{align}
The standard error of $\hat{\mu}$ can then be estimated as $m^{-1/2} \hat{\sigma}$.

\newpage
\section{Explicitly writing down the symmetry of the cat/dog game} \label{appendix:cat/dog}

We first specify the cat/dog game in Figure \ref{fig:XP_cat/dog} as a Dec-POMDP, and then show that it contains a non-trivial Dec-POMDP symmetry which corresponds to the fact that one can swap the two pets.

We write $\mathcal{S} = \cup_{t = 0}^2 \mathcal{S}_{t}$, where
\begin{align}
\begin{split}
    s_0 \in \mathcal{S}_0 :=& \{ \text{cat}, \text{dog} \}, \\
    s_1 \in \mathcal{S}_1 :=& \{ (\text{cat}, \text{bail}), (\text{cat}, \text{reveal}), (\text{cat}, \text{on}), (\text{cat}, \text{off}), (\text{dog}, \text{bail}), (\text{dog}, \text{reveal}), (\text{dog}, \text{on}), (\text{dog}, \text{off}) \}, \\
    s_2 \in \mathcal{S}_2 :=& \{ \text{terminal} \}.
\end{split}
\end{align}
The observation functions are then given as follows:
\begin{align}
\begin{split}
    \mathcal{U}^1(s_0) =& s_0, \\
    \mathcal{U}^2(s_0) =& \text{dummy}, \\
    \mathcal{U}^1(s_1) =& \text{dummy}, \\
    \mathcal{U}^2((\text{cat/dog}, \text{bail})) =& \text{dummy}, \\
    \mathcal{U}^2((\text{cat}, \text{reveal})) =& \text{cat}, \\
    \mathcal{U}^2((\text{dog}, \text{reveal})) =& \text{dog}, \\
    \mathcal{U}^2((\text{cat/dog}, \text{on})) =& \text{on}, \\
    \mathcal{U}^2((\text{cat/dog}, \text{off})) =& \text{off}.
\end{split}
\end{align}
The local action spaces are then given as follows:
\begin{align}
\begin{split}
    \mathcal{A}^1(o_1^1 = \text{dummy}) =& \{ \text{dummy} \}, \\
    \mathcal{A}^2(o_0^2 = \text{dummy}) =& \{ \text{dummy} \}, \\
    \mathcal{A}^2(o_1^2 = \text{dummy}) =& \{ \text{dummy} \}, \\
    \mathcal{A}^1(o_0^1 = \text{cat/dog}) =& \{ \text{bail}, \text{reveal}, \text{on}, \text{off} \}, \\
    \mathcal{A}^2(o_1^2 = \text{cat/dog/on/off}) =& \{ \text{bail}, \text{cat}, \text{dog} \}.
\end{split}
\end{align}
The initial state distribution is uniform on $\mathcal{S}_0$, and the deterministic state transition function is obtained in the obvious manner: choosing joint action $a_0$ in state $s_0$ leads to the state $s_1$ which is given by concatenating $s_0$ and the non-dummy part of $a_0$. The state $s_2$ is always terminal.

The reward function is given as follows:
\begin{align}
\begin{split}
    \mathcal{R}( s_0 = \text{cat/dog}, a_0 = \text{reveal}) =& -3, \\
    \mathcal{R}( s_0 = \text{cat/dog}, a_0 = \text{bail}) =& 1, \\
    \mathcal{R}( s_0 = \text{cat/dog}, a_0 = \text{on/off}) =& 0, \\
    \mathcal{R}( s_1 = (\text{cat/dog}, \text{bail}), a_1 = \text{dummy}) =& 0, \\
    \mathcal{R}( s_1 = (\text{cat/dog}, \text{reveal/on/off}), a_1 = \text{cat/dog}) =& 10, \\
    \mathcal{R}( s_1 = (\text{cat/dog}, \text{reveal/on/off}), a_1 = \text{dog/cat}) =& -10, \\
    \mathcal{R}( s_1 = (\text{cat/dog}, \text{reveal/on/off}), a_1 = \text{bail}) =& 1. \\
\end{split}
\end{align}
The non-trivial Dec-POMDP symmetry $\phi$ is then given as follows:
\begin{align}
\begin{split}
    \phi_{\mathcal{S}}(\text{cat/dog}) =& \text{dog/cat}, \\
    \phi_{\mathcal{S}}((\text{cat/dog}, \text{bail/reveal/on/off})) =& (\text{dog/cat}, \text{bail/reveal/on/off}), \\
    \phi_{\mathcal{O}^1}(o_0^1 = \text{cat/dog}) =& \text{dog/cat}, \\
    \phi_{\mathcal{O}^2}(o_1^2 = \text{cat/dog}) =& \text{dog/cat}, \\
    \phi_{\mathcal{O}^2}(o_1^2 = \text{on/off}) =& \text{on/off}, \\
    \phi_{\mathcal{A}^1}(a_0^1) =& a_0^1, \\
    \phi_{\mathcal{A}^2}(a_1^2 = \text{cat/dog}) =& \text{dog/cat}, \\
    \phi_{\mathcal{A}^2}(a_1^2 = \text{bail}) =& \text{bail}.
\end{split}
\end{align}
It is now straightforward to check that $\phi$ is a Dec-POMDP symmetry. 

Applying $\phi$ to a joint policy $\pi$, we see that
\begin{align}
\begin{split}
    \phi(\pi^1)(\text{on/off/bail/reveal} | \text{cat/dog}) =& \pi^1(\text{on/off/bail/reveal} | \text{dog/cat}), \\
    \phi(\pi^2)(\text{cat/dog} | \text{on/off}) =& \pi^2(\text{dog/cat} | \text{on/off}), \\
    \phi(\pi^2)(\text{cat} | \text{cat}) =& \pi^2(\text{dog} | \text{dog}), \\
    \phi(\pi^2)(\text{dog} | \text{dog}) =& \pi^2(\text{cat} | \text{cat}).
\end{split}
\end{align}
Therefore, if $\pi$ is symmetry-equivariant, i.e. $\phi(\pi) = \pi$, we see that none of the two signaling conventions can be preferred. Thus among the symmetry-equivariant policies, the policy under which Alice always reveals and under which Bob always guesses the pet he saw, is optimal.

\section{Another Toy Game in which one cannot find the optimal symmetry-equivariant policy through high entropy regularization} \label{appendix:second toygame}

Consider the one-round simultaneous-action game with payoff matrix
\begin{align}
    \begin{bmatrix}
        3 & 0 & 0 \\
        0 & 3 & 0 \\
        0 & 0 & 2
    \end{bmatrix}.
\end{align}
For high enough $\alpha$, the local policies of the unique $\pi_{\theta^\alpha}$ will assign equal probability to the first two actions, and a lower probability to the third one. The greedification $\hat{\pi}_{\theta^\alpha}$ will have a return of $1.5$, while the optimal symmetry-equivariant policy, under which both local policies always choose the third action, has a return of $2$. Intuitively, this means that a high $\alpha$ will ensure that the learned policies do not break the symmetries of the Dec-POMDP, but not necessarily that they \textit{exploit} the symmetries.

\newpage
\section{Biased Advantage Estimates can also lead to Symmetry-Breaking} \label{appendix:GAE}

When using a critic and generalized advantage estimates (GAE) \cite{schulman2015high} in order to reduce the variance in the estimates of $\nabla_\theta^\alpha J_{\text{SP}}$, we are adding bias to those estimates, since the critic will not be fully accurate. This bias might also lead to symmetry-breaking, since an asymmetrically biased critic will lead the actor to break symmetries. As a simple demonstration of this, we ran IPPO with $\alpha = 0.01, 0.05, 0.10$, and different $\lambda_{\text{GAE}}$, see Table \ref{table:GAE}. Note that for all the Hanabi results in the main paper we used $\lambda_{\text{GAE}} = 0.9$. We see that when $\lambda_{\text{GAE}}$ is too small, i.e. when the GAEs heavily rely on the critic and thus have low variance but high bias, then this results in a gap between SP and XP. When $\lambda_{\text{GAE}}$ is too large though, then this can lead to both SP and XP decreasing.

\begin{table}[ht]
\centering
\caption{2-Player Hanabi: Mean $\pm$ standard error of greedy SP and XP, estimated with (\ref{eq:mean XP naive}) and (\ref{eq:sigma XP naive}), of IPPO LSTM policies trained with different $\alpha$ and $\lambda_{\text{GAE}}$. 4 seeds per set of hyperparameters.}
\label{table:GAE}
\begin{tabular}{c|c|c|c}
\hline
$\alpha$ & $\lambda_{\text{GAE}}$ & SP & XP \\
\hline
0.01 & 0.00 & $22.75 \pm 0.41$ & $18.04 \pm 1.01$ \\
0.01 & 0.50 & $\textbf{24.43} \pm 0.08$ & $5.58 \pm 8.08$ \\
0.01 & 0.80 & $24.41 \pm 0.07$ & $14.36 \pm 6.45$ \\
0.01 & 0.90 & $24.18 \pm 0.10$ & $22.78 \pm 0.79$ \\
0.01 & 0.95 & $24.16 \pm 0.01$ & $22.59 \pm 1.00$ \\
0.01 & 1.00 & $23.52 \pm 0.08$ & $\textbf{23.14} \pm 0.25$ \\
\hline
0.05 & 0.00 & $22.45 \pm 0.20$ & $20.09 \pm 1.42$ \\
0.05 & 0.50 & $24.43 \pm 0.03$ & $7.940 \pm 9.60$ \\
0.05 & 0.80 & $\textbf{24.51} \pm 0.01$ & $16.18 \pm 7.37$ \\
0.05 & 0.90 & $24.47 \pm 0.01$ & $\textbf{24.48} \pm 0.02$ \\
0.05 & 0.95 & $24.31 \pm 0.03$ & $24.23 \pm 0.08$ \\
0.05 & 1.00 & $14.96 \pm 0.01$ & $14.96 \pm 0.01$ \\
\hline
0.10 & 0.00 & $22.12 \pm 0.17$ & $20.99 \pm 1.03$ \\
0.10 & 0.50 & $24.16 \pm 0.19$ & $22.13 \pm 2.08$ \\
0.10 & 0.80 & $\textbf{24.35} \pm 0.02$ & $\textbf{24.32} \pm 0.03$ \\
0.10 & 0.90 & $24.14 \pm 0.02$ & $24.13 \pm 0.02$ \\
0.10 & 0.95 & $14.99 \pm 0.01$ & $14.99 \pm 0.01$ \\
0.10 & 1.00 & $10.07 \pm 0.02$ & $10.07 \pm 0.01$ \\
\hline
\end{tabular}
\end{table}

\newpage

\section{Hyperparameters} \label{appendix:hyperparameters}

\begin{figure}
\centering
\begin{tikzpicture}

\node[box] (lstm) {LSTM};
\node[box, left=15mm of lstm] (mlp1) {MLP};
\node[box, right=15mm of lstm] (mlp2) {MLP};

\node[box, align=center, below=6mm of mlp1] (private) {\small Private Observation $o_t^{i, \text{private}}$};
\draw[->] (private) -- (mlp1.south);
\node[box, align=center, below=6mm of lstm] (public) {\small Public Observation $o_t^{i, \text{public}}$};
\draw[->] (public) -- (lstm.south);
\node[box, align=center, below=6mm of mlp2] (state) {\small State $s_t$};
\draw[->] (state) -- (mlp2.south);

\node[below left=-4mm and 3mm of lstm] (htlab1) {$h_{t-1}^i$};
\draw[->] (htlab1) -- (lstm.west);
\node[above right=-4mm and 3mm of lstm] (htlab2) {$h_{t}^i$};
\draw[->] (lstm.east) -- (htlab2);

\node[dotprod, above left=10mm and 8mm of lstm] (times1) {\small $\times$};
\node[dotprod, above right=10mm and 8mm of lstm] (times2) {\small $\times$};
\draw[->] (lstm.north) -- (times1.east);
\draw[->] (mlp1.north)  -- (times1.west);
\draw[->] (lstm.north) -- (times2.west);
\draw[->] (mlp2.north)  -- (times2.east);

\node[box, above=8mm of times1] (actor) {};
\node[align=center] at (actor) {Actor $\pi_\theta^i(\cdot | \tau_t^i)$};
\draw[->] (times1) -- (actor);

\node[box, above=8mm of times2] (critic) {};
\node[align=center] at (critic) {Critic $V_{\pi_\theta}^\varphi(\tau_t)$};
\draw[->] (times2) -- (critic);

\end{tikzpicture}

\caption{Public-Private LSTM Actor-Critic Architecture: In Hanabi, a player's public observation includes everything except for the player hands. A player's private observation is the public observation plus the other players' hands. The state is the public observation plus all players' hands. For IPPO one technically does not need a separate MLP for the critic, as the critic conditions only the local AOH $\tau_t^i$, just like the actor, but we still use a critic MLP which just receives the private observation $o_t^{i, \text{private}}$ as well. Both MLP streams have 3 hidden layers, and the LSTM stream has one feedforward embedding layer and two LSTM layers.}
\label{fig:public-private LSTM}
\end{figure}

\subsection{Hanabi}

In all architectures we use a width of $512$ in all layers. All non-LSTM layers have a ReLU activation. For a description of the public-private LSTM architecture see Figure \ref{fig:public-private LSTM}. Furthermore we share weights between all agents.

For the PPO hyperparameters that are constant across all our Hanabi experiments, see Table \ref{table:ppo_hyperparams_hanabi} below. The only non-standard implementation detail in our code is that we mask out the actor loss and the entropy bonus for the non-acting player, as it always contributes zero to the gradient. This prevents the actor loss and the entropy bonus from effectively being divided by the number of agents and allows for more consistent hyperparameters across 2, 3, 4, and 5 player Hanabi.

To train our OBL agents, we used $10^{10}$ total timesteps to train level 1, and $5 \times 10^{9}$ to train all subsequent levels, where we start training of level $k$ with the final weights from level $k-1$. The architecture of the belief model is exactly the same as the one used in \cite{hu2021off}, and is trained through supervised learning.

\begin{table}[h!]
\centering
\caption{PPO hyperparameters that are fixed across all our Hanabi experiments.}
\label{table:ppo_hyperparams_hanabi}
\begin{tabular}{l c}
\toprule
\textbf{Hyperparameter} & \textbf{Value} \\
\midrule
Learning Rate                 & $5\times 10^{-4}$ \\
Number of Environments        & $1024$ \\
Number of Steps per Rollout   & $128$ \\
Total Timesteps               & $10^{10}$ \\
Update Epochs                 & $4$ \\
Number of Minibatches         & $4$ \\
Discount Factor ($\gamma$)    & $0.999$ \\
Clipping Coefficient          & $0.2$ \\
Value Function Coefficient    & $0.5$ \\
Max Gradient Norm             & $0.5$ \\
Linear Learning Rate Annealing& True \\
Optimizer                     & Adam \\
Initialization                & Orthogonal \\
\bottomrule
\end{tabular}
\end{table}

\subsection{Overcooked}

Our actor-critic network for Overcooked is shared between both agents and has the following architecture: the observations are embedded through 3 CNN layers with 32 features each, and kernel sizes of $(5, 5)$, $(3, 3)$, $(3, 3)$, with ReLU activations in between, followed by a fully connected layer with output dimension 64. This embedding is the concatenated with a one-layer fully connected embedding of the environment timestep $t$. This concatenation is fed through two further fully connected layers with output dimension 64 and ReLU activation, after which there are separate actor and critic heads. We gave the environment timestep to the actor-critic network since otherwise the input is not Markov as the agents cannot tell when the game ends. The PPO hyperparameters are given in Table \ref{table:ppo_hyperparams_overcooked}
\begin{table}[h!]
\centering
\caption{PPO hyperparameters that are fixed across all our Overcooked experiments.}
\label{table:ppo_hyperparams_overcooked}
\begin{tabular}{l c}
\toprule
\textbf{Hyperparameter} & \textbf{Value} \\
\midrule
Learning Rate                 & $4\times 10^{-4}$ \\
Number of Environments        & $64$ \\
Number of Steps per Rollout   & $256$ \\
Total Timesteps               & $10^{8}$ \\
Reward Shaping Timesteps      & $5 \times 10^{7}$ \\
Update Epochs                 & $4$ \\
Number of Minibatches         & $16$ \\
Discount Factor ($\gamma$)    & $0.99$ \\
$\lambda_{\text{GAE}}$        & $0.8$ \\
Clipping Coefficient          & $0.2$ \\
Value Function Coefficient    & $0.5$ \\
Max Gradient Norm             & $0.5$ \\
Linear Learning Rate Annealing& True \\
Optimizer                     & Adam \\
Initialization                & Orthogonal \\
\bottomrule
\end{tabular}
\end{table}

\newpage
\subsection{Yokai}
Our actor-critic network for Yokai and our chosen hyperparameters are adapted from  \cite{ruhdorfer2025yokailearningenvironmenttracking}.
Specifically, we use their best-performing CNN encoder which uses 4 CNN layers with $64$ filters, kernel size $(3,3)$ and ReLU activation each.
The stride is 1 and we employ valid padding.
The resulting embedding is fed to a linear projection (to project the CNN output to the GRU hidden size), followed by a GRU, then four feedforward layers, and finally to the actor and critic heads, respectively.
The hidden dimension of the dense layers and the GRU is 256.
All hyperparameters are shown in Table \ref{table:ppo_hyperparams_yokai}.

All our experiments were conducted in the 3x3 version of Yokai with memory help turned on.
This corresponds to the ZSC experimental setting in the original paper.
For all environment details, we refer the reader to the original work \cite{ruhdorfer2025yokailearningenvironmenttracking}.
\begin{table}[h!]
\centering
\caption{PPO hyperparameters that are fixed across all our Yokai experiments.}
\label{table:ppo_hyperparams_yokai}
\begin{tabular}{l c}
\toprule
\textbf{Hyperparameter} & \textbf{Value} \\
\midrule
Learning Rate                 & $3 \times 10^{-4}$ \\
Number of Environments        & $1024$ \\
Number of Steps per Rollout   & $128$ \\
Total Timesteps               & $10^9$ \\
Reward Shaping Timesteps      & $10^9$ (linearly annealed) \\
Update Epochs                 & $4$ \\
Number of Minibatches         & $4$ \\
Discount Factor ($\gamma$)    & $0.99$ \\
$\lambda_{\text{GAE}}$        & $0.85$ \\
Clipping Coefficient          & $0.2$ \\
Value Function Coefficient    & $0.5$ \\
Max Gradient Norm             & $0.5$ \\
Linear Learning Rate Annealing& True \\
Optimizer                     & Adam \\
Initialization                & Orthogonal \\
\bottomrule
\end{tabular}
\end{table}

\newpage
\section{Entropy threshold in the Overcooked layout ``cramped room''} \label{appendix:cramped room}

\begin{figure}[h]
    \begin{center}
        \includegraphics[scale=0.4]{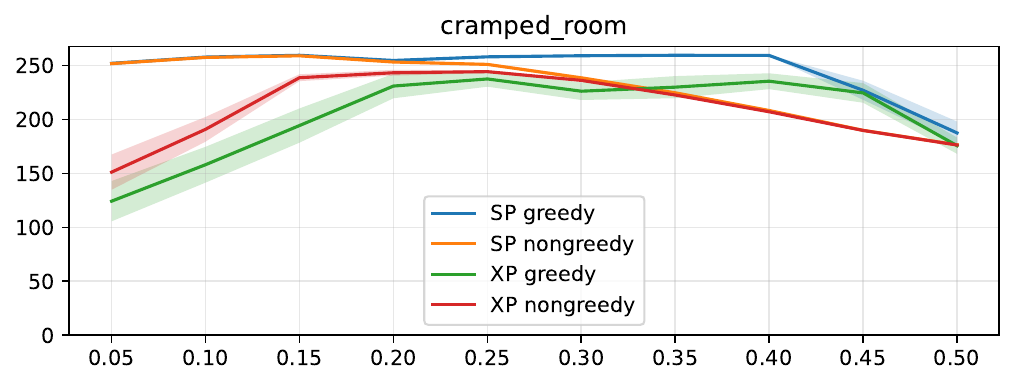}
    \end{center}
    \caption{The standard Overcooked layout ``cramped room'': Mean SP and XP, greedy and non-greedy, of IPPO with different entropy coefficients $\alpha$. The shaded regions show one standard error of the mean, estimated with (\ref{eq:mean XP correct}) and (\ref{eq:sigma XP correct}). We used $\lambda_{\text{GAE}} = 0.85$ and $16$ seeds per $\alpha$. \textbf{Takeaway}: As we increase $\alpha$, non-greedy XP approaches non-greedy SP, greedy XP approaches greedy SP, and greedy SP/XP are higher than non-greedy SP/XP.}
\end{figure}

\newpage
\section{Full XP Matrices in Hanabi} \label{appendix:full XP matrices}

\begin{figure}[H]
    \begin{center}
        \includegraphics[scale=0.6]{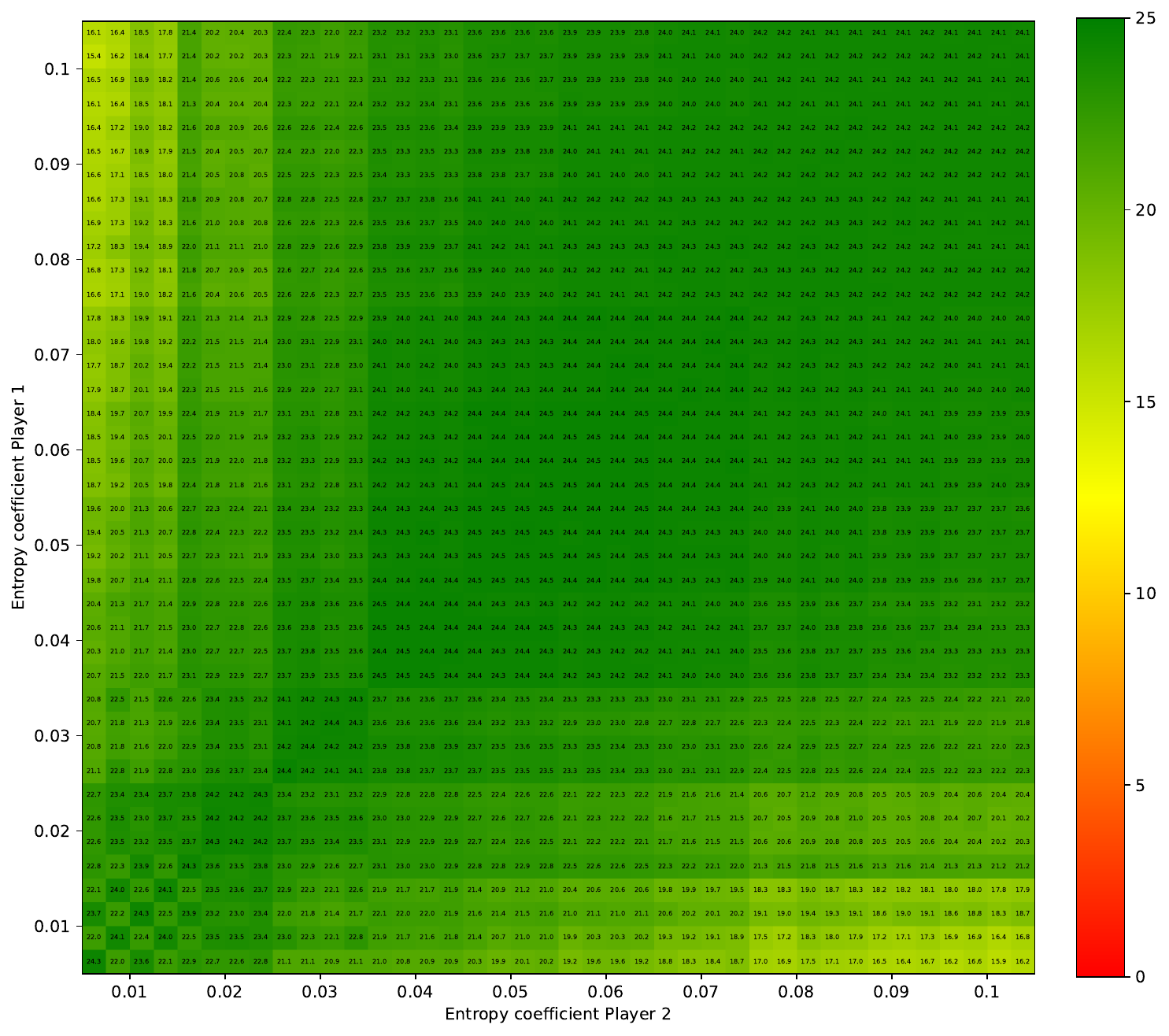}
    \end{center}
    \caption{2-Player Hanabi: XP matrix between LSTM IPPO policies trained with different entropy coefficients. Four seeds per entropy coefficient $0.01, 0.02, ..., 0.10$.}
    \label{fig:XP_IPPO_LSTM_FULL}
\end{figure}

\begin{figure}[H]
    \begin{center}
        \includegraphics[scale=0.6]{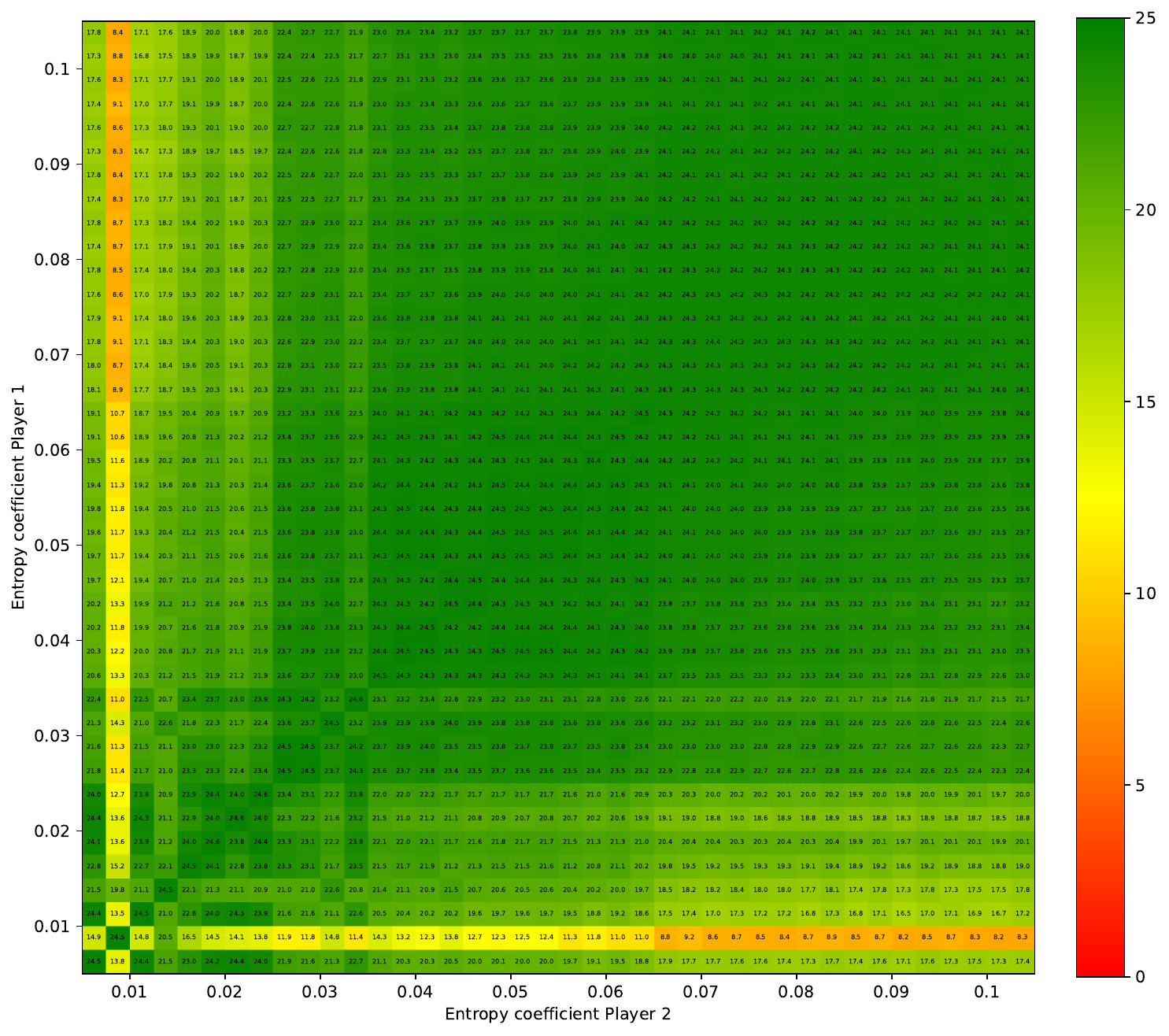}
    \end{center}
    \caption{2-Player Hanabi: XP matrix between public-private LSTM IPPO policies trained with different entropy coefficients. Four seeds per entropy coefficient $0.01, 0.02, ..., 0.10$.}
    \label{fig:XP_IPPO_LSTM_PP_FULL}
\end{figure}

\begin{figure}[H]
    \begin{center}
        \includegraphics[scale=0.6]{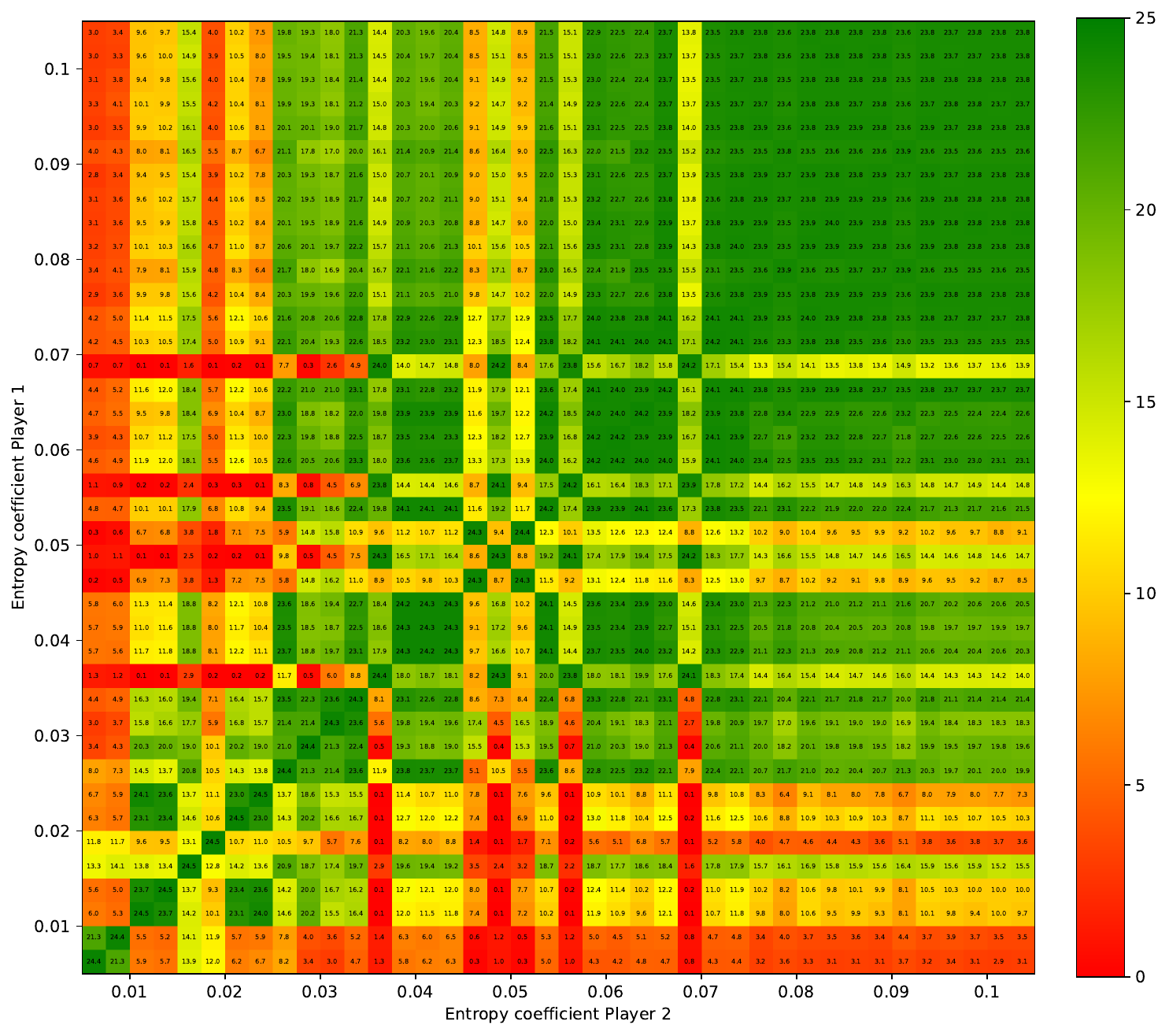}
    \end{center}
    \caption{2-Player Hanabi: XP matrix between feedforward IPPO policies trained with different entropy coefficients. Four seeds per entropy coefficient $0.01, 0.02, ..., 0.10$.}
    \label{fig:XP_IPPO_FF_FULL}
\end{figure}

\section{Block XP Matrices in Overcooked} \label{appendix:Overcooked Block XP}

\begin{figure}[H]
    \begin{center}
        \includegraphics[scale=0.4]{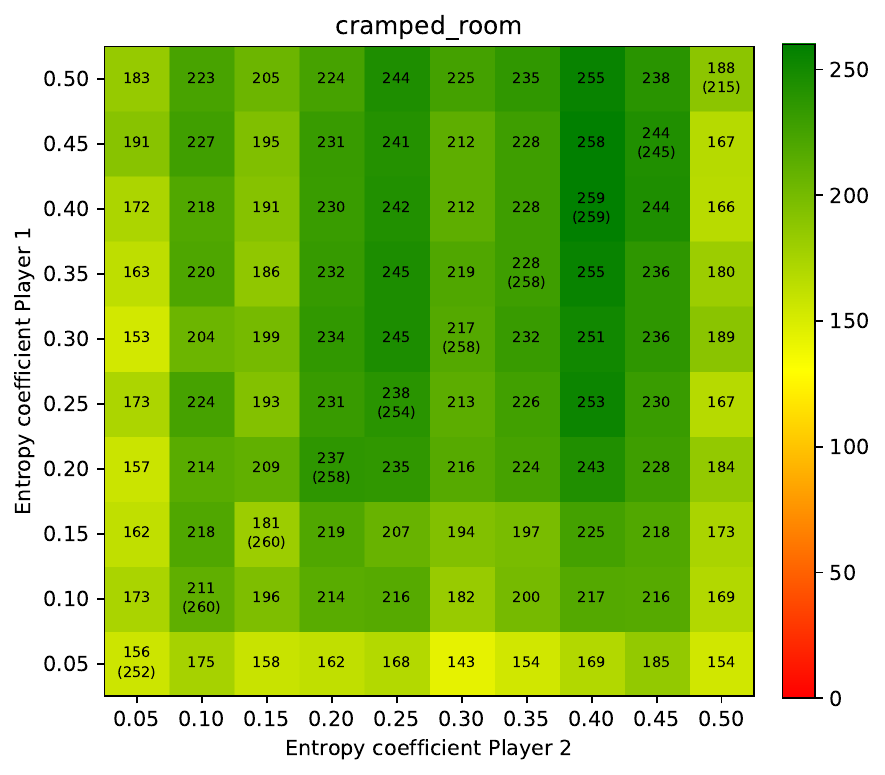}
        \includegraphics[scale=0.4]{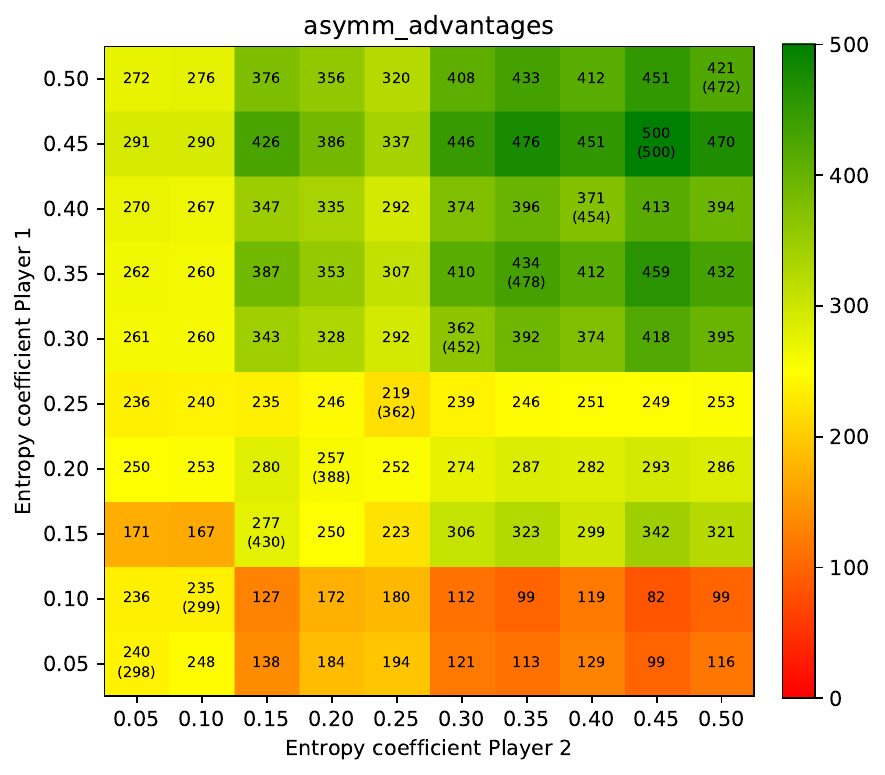}
        \includegraphics[scale=0.4]{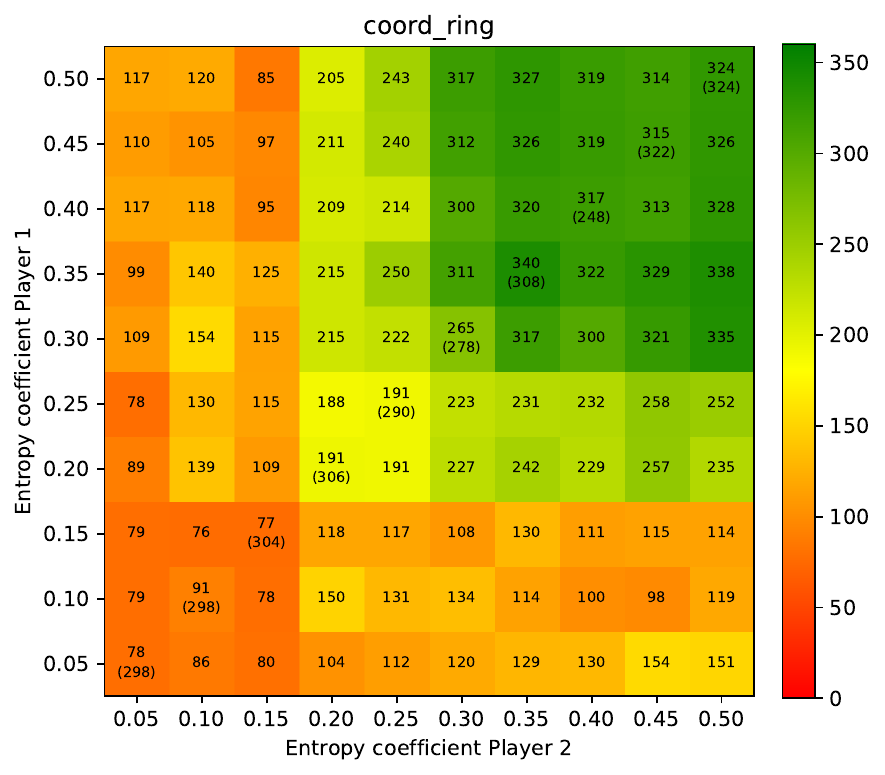}
        \includegraphics[scale=0.4]{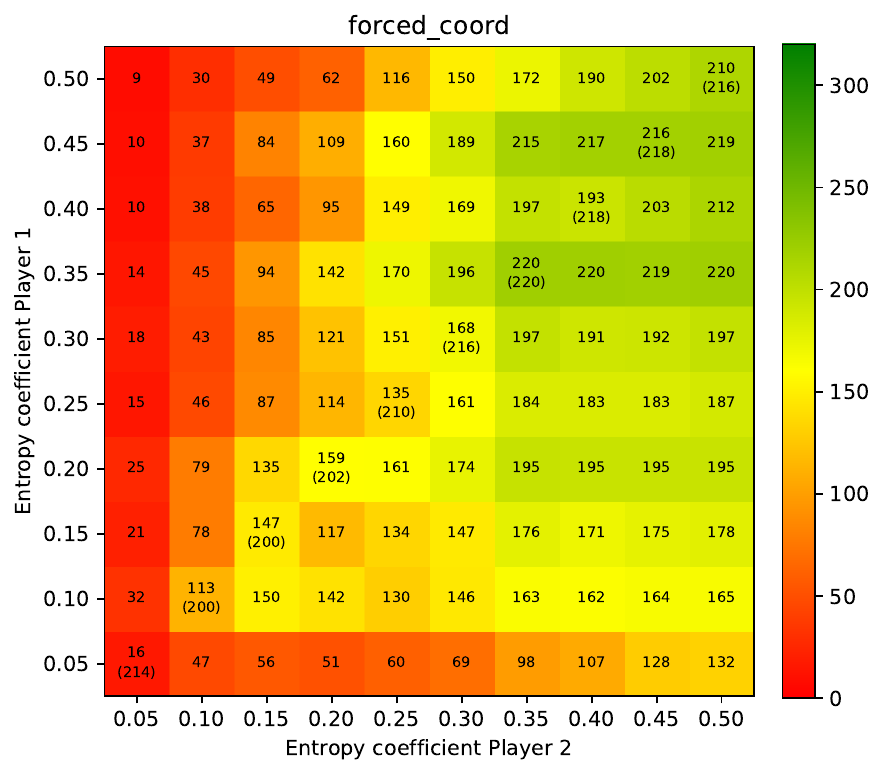}
        \includegraphics[scale=0.4]{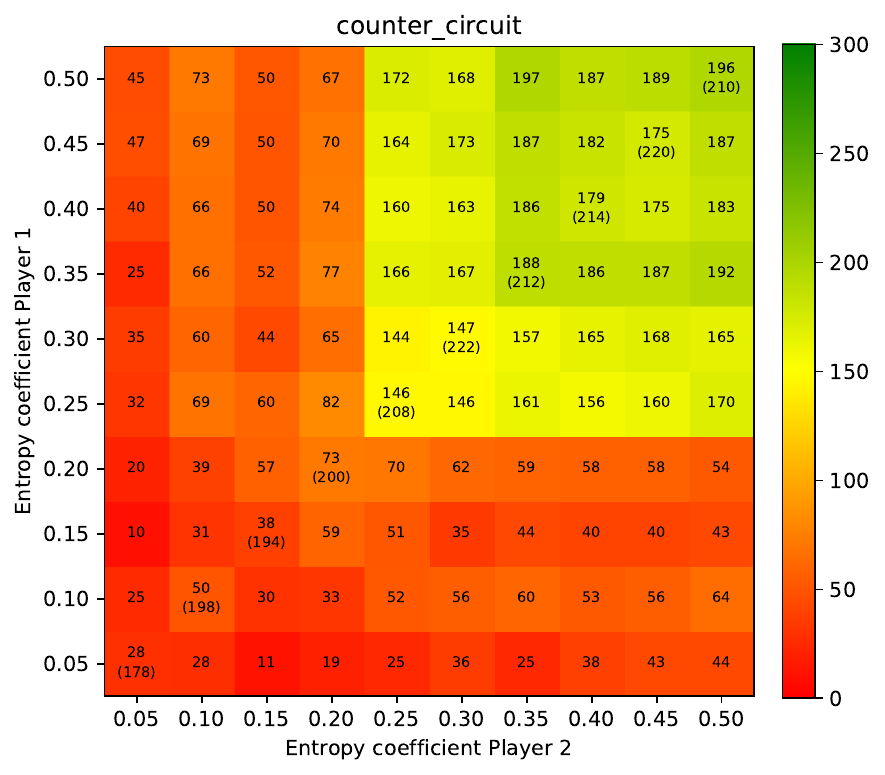}
    \end{center}
    \caption{Overcooked: Block XP matrices for 100 greedified IPPO policies, for different entropy coefficients $\alpha$, and $\lambda_{\text{GAE}} = 0.8$. Ten seeds per $\alpha$. On the diagonal, the numbers are the average SP (in parentheses) and the average of the off-diagonal in the diagonal 10x10 blocks in which all the policies are trained with the same $\alpha$. The off-diagonal numbers are the average of the off-diagonal 10x10 blocks in which policies come from different $\alpha$.}
    \label{fig:XP_Overcooked}
\end{figure}

\newpage

\end{document}